\documentclass{article}

\usepackage{authblk}
\usepackage[utf8]{inputenc}
\usepackage{main} 
\usepackage{microtype}
\usepackage{times}
\usepackage{latexsym}
\usepackage{algorithm}
\usepackage{algorithmic}
\usepackage{graphicx}
\usepackage{xcolor}
\usepackage{wrapfig}
\definecolor{deepgreen}{RGB}{34,139,34}
\usepackage{mdframed}

\usepackage{tcolorbox}
\usepackage{verbatim}
\usepackage{listings}
\lstnewenvironment{TextBlock}{
  \lstset{
    basicstyle=\small\ttfamily,
    breaklines=true,
    breakatwhitespace=false,
    columns=flexible
  }
}{}

\usepackage{tikz}
\usetikzlibrary{arrows.meta, positioning, fit, backgrounds, decorations.pathreplacing}
\newtcolorbox{CompareBox}[2][]{%
  enhanced, breakable,
  colback=#2!5!white,
  colframe=#2!50!black,
  boxrule=1pt,
  arc=6pt,
  top=6pt, bottom=6pt, left=6pt, right=6pt,
  title=#1,
  fonttitle=\bfseries,
}
\tcbuselibrary{breakable,skins}
\usepackage{longtable}

\usepackage[table]{xcolor}
\usepackage{colortbl}  

\usepackage{amsmath}
\usepackage{amssymb}
\usepackage{amsfonts}
\usepackage{mathtools}
\usepackage{bm}

\usepackage{graphicx}
\usepackage{subcaption}
\usepackage{float}
\usepackage{wrapfig}

\usepackage{booktabs}
\usepackage{array}
\usepackage{multirow}
\usepackage{multicol}
\usepackage{makecell}
\usepackage{longtable}
\usepackage{tabularx}
\usepackage{colortbl}
\usepackage{arydshln}

\usepackage{color}
\usepackage{xcolor}
\definecolor{mydarkblue}{rgb}{0,0.08,0.45}
\definecolor{wkblue}{rgb}{0.2, 0.3, 0.6}
\definecolor{meta-color}{rgb}{0.5, 0.5, 0.5}
\definecolor{bgblue}{RGB}{245,243,253}
\definecolor{ttblue}{RGB}{91,194,224}
\definecolor{mybrown}{RGB}{128,64,0}
\definecolor{titlecolor}{HTML}{4c9cff}
\definecolor{coolblue3}{rgb}{0.91, 0.94, 0.98}
\definecolor{myblue}{rgb}{0.9, 0.1, 0.94}
\definecolor{mygreen}{rgb}{0.64, 0.56, 0.88}
\definecolor{myyellow}{rgb}{0.68, 0.6, 0.1}
\definecolor{fancygreen}{rgb}{0.33, 0.68, 0.20}
\definecolor{salmon}{rgb}{0.94, 0.52, 0.49}
\definecolor{tablegreen}{rgb}{0.82, 0.94, 0.75}
\definecolor{tableblue}{rgb}{0.81, 0.90, 0.94}
\definecolor{tablered}{rgb}{0.97, 0.85, 0.85}
\definecolor{tableorange}{rgb}{0.96, 0.85, 0.81}

\usepackage{enumitem}
\usepackage{footnote}
\usepackage{lipsum}
\usepackage{setspace}

\usepackage{geometry}
\usepackage{fancyhdr}
\usepackage{lscape}
\usepackage{rotating}

\usepackage{algorithm}
\usepackage{algorithmic}

\usepackage{awesomebox}

\usepackage{natbib}
\usepackage{url}
\usepackage[colorlinks=true,linkcolor=mydarkblue,citecolor=mydarkblue,filecolor=mydarkblue,urlcolor=mydarkblue]{hyperref}

\usepackage{bbding}
\usepackage{imakeidx}
\makeindex
\usepackage[toc]{multitoc}
\usepackage[edges]{forest}
\usepackage[normalem]{ulem}
\usepackage{fontawesome5}
\usepackage{blindtext}
\usepackage{pgfplots}
\usepackage{pgfplotstable}

\usepackage{listings}
\usepackage{listingsutf8}
\newcommand\JSONnumbervaluestyle{\color{blue}}
\newcommand\JSONstringvaluestyle{\color{red}}

\newif\ifcolonfoundonthisline

\makeatletter
\lstdefinestyle{json}
{
  showstringspaces    = false,
  keywords            = {false,true},
  alsoletter          = 0123456789.,
  morestring          = [s]{"}{"},
  stringstyle         = \ifcolonfoundonthisline\JSONstringvaluestyle\fi,
  MoreSelectCharTable =%
    \lst@DefSaveDef{`:}\colon@json{\processColon@json},
  basicstyle          = \ttfamily,
  keywordstyle        = \ttfamily\bfseries,
}

\newcommand\processColon@json{%
  \colon@json%
  \ifnum\lst@mode=\lst@Pmode%
    \global\colonfoundonthislinetrue%
  \fi
}

\lst@AddToHook{Output}{%
  \ifcolonfoundonthisline%
    \ifnum\lst@mode=\lst@Pmode%
      \def\lst@thestyle{\JSONnumbervaluestyle}%
    \fi
  \fi
  \lsthk@DetectKeywords%
}

\lst@AddToHook{EOL}%
  {\global\colonfoundonthislinefalse}
\makeatother

\newtcolorbox{myboxi}[1][]{
  breakable,
  title=#1,
  colback=red!5,
  colbacktitle=red!5,
  coltitle=black,
  fonttitle=\bfseries,
  bottomrule=0pt,
  toprule=0pt,
  leftrule=2pt,
  rightrule=2pt,
  titlerule=0pt,
  arc=0pt,
  outer arc=0pt,
  colframe=red,
}

\newtcolorbox{myboxnote}[1][]{
  breakable,
  title=#1,
  colback=orange!0,
  colbacktitle=orange!0,
  coltitle=black,
  fonttitle=\bfseries,
  bottomrule=0pt,
  toprule=0pt,
  leftrule=2pt,
  rightrule=2pt,
  titlerule=0pt,
  arc=0pt,
  outer arc=0pt,
  colframe=orange,
}

\newtcolorbox{myboxii}[1][]{
  breakable,
  freelance,
  title=#1,
  colback=white,
  colbacktitle=white,
  coltitle=black,
  fonttitle=\bfseries,
  bottomrule=0pt,
  boxrule=0pt,
  colframe=white,
  overlay unbroken and first={
  \draw[red!75!black,line width=3pt]
    ([xshift=5pt]frame.north west) --
    (frame.north west) --
    (frame.south west);
  \draw[red!75!black,line width=3pt]
    ([xshift=-5pt]frame.north east) --
    (frame.north east) --
    (frame.south east);
  },
  overlay unbroken app={
  \draw[red!75!black,line width=3pt,line cap=rect]
    (frame.south west) --
    ([xshift=5pt]frame.south west);
  \draw[red!75!black,line width=3pt,line cap=rect]
    (frame.south east) --
    ([xshift=-5pt]frame.south east);
  },
  overlay middle and last={
  \draw[red!75!black,line width=3pt]
    (frame.north west) --
    (frame.south west);
  \draw[red!75!black,line width=3pt]
    (frame.north east) --
    (frame.south east);
  },
  overlay last app={
  \draw[red!75!black,line width=3pt,line cap=rect]
    (frame.south west) --
    ([xshift=5pt]frame.south west);
  \draw[red!75!black,line width=3pt,line cap=rect]
    (frame.south east) --
    ([xshift=-5pt]frame.south east);
  },
}

\mdfdefinestyle{mystyle}{%
  rightline=true,
  innerleftmargin=10,
  innerrightmargin=10,
  outerlinewidth=3pt,
  topline=false,
  rightline=true,
  bottomline=false,
  skipabove=\topsep,
  skipbelow=\topsep
}

\usepackage{pifont}

\DeclareCaptionFont{black}{\color{black}}

\newenvironment{itemize*}%
 {\leftmargini=10pt\begin{itemize}%
  \setlength{\itemsep}{0pt}%
  \setlength{\parskip}{0pt}%
  }%
 {\end{itemize}}
\newenvironment{enumerate*}%
 {\begin{enumerate}%
  \setlength{\itemsep}{0pt}%
  \setlength{\parskip}{0pt}}%
 {\end{enumerate}}

\usepackage{etoolbox}
\newcounter{bibcount}
\makeatletter
\patchcmd{\@lbibitem}{\item[}{\item[\hfil\stepcounter{bibcount}{[\thebibcount]}}{}{}
\renewcommand\NAT@bibsetup%
  [1]{\setlength{\leftmargin}{\bibhang}\setlength{\itemindent}{-\parindent}%
      \setlength{\itemsep}{\bibsep}\setlength{\parsep}{\z@}}
\makeatother

\definecolor{myblue}{rgb}{0.9, 0.1, 0.94}
\definecolor{mygreen}{rgb}{0.64, 0.56, 0.88}
\definecolor{myyellow}{rgb}{0.68, 0.6, 0.1}
\definecolor{fancygreen}{rgb}{0.33, 0.68, 0.20}
\definecolor{salmon}{rgb}{0.94, 0.52, 0.49}
\definecolor{tablegreen}{rgb}{0.82, 0.94, 0.75}
\definecolor{tableblue}{rgb}{0.81, 0.90, 0.94}
\definecolor{tablered}{rgb}{0.97, 0.85, 0.85}
\definecolor{tableorange}{rgb}{0.96, 0.85, 0.81}

\DeclareMathOperator*{\argmax}{arg\,max}

\begin{document}


\title{AutoCurator: Automating Data Curation Strategy Design at Pretraining Scale}

\title{Data Darwinism -- Part II\\ Automating Data Curation Strategy Design at Pretraining Scale}

\title{Data Darwinism -- Part II: \modelname \\ Evolving Curation Strategies at Pretraining Scale}

\title{\modelname: AI can Discover Better Pretraining Data}

\title{Data Darwinism -- Part II: \modelname \\ AI Evolves Better Pretraining Data Curation Strategies}

\title{\modelname: AI Autonomously Evolves Pretraining Data Curation}
\title{AI can Autonomously Evolve Pretraining Data Curation}

\author[1,2,4]{Tiantian Mi}
\author[5]{Dongming Shan}
\author[1,2,4]{Zhen Huang}
\author[1,3,4]{Yiwei Qin}
\author[1,4]{Muhang Xie}
\author[4]{\authorcr Yuxuan Qiao}
\author[1,3,4]{Yixiu Liu}
\author[1,3,4]{Chenyang Zhou}
\author[1,3,4]{Pengfei Liu\textsuperscript{†}}
\affil[ ]{\textsuperscript{1}SII \quad \textsuperscript{2}FDU \quad \textsuperscript{3}SJTU \quad \textsuperscript{4}GAIR \quad \textsuperscript{5}KPS}

\newcommand{\modelname}{\textit{DataEvolve}\xspace}
\newcommand{\dataname}{\texttt{Darwin-CC}\xspace}

\maketitle


\pagestyle{fancy}
\thispagestyle{fancy}
\fancyhead{}
\lhead{
  \raisebox{-0.3cm}{\includegraphics[height=0.95cm]{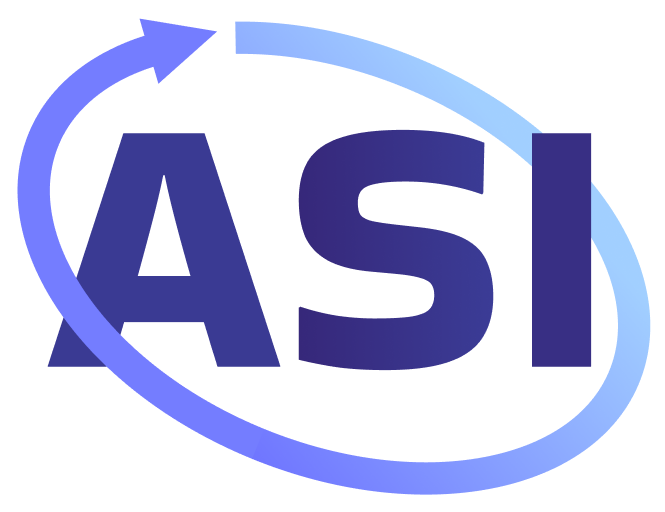}}
}
\chead{Data Darwinism -- Part II: \modelname}
\rhead{%
  \raisebox{-0.2cm}{\includegraphics[height=0.7cm]{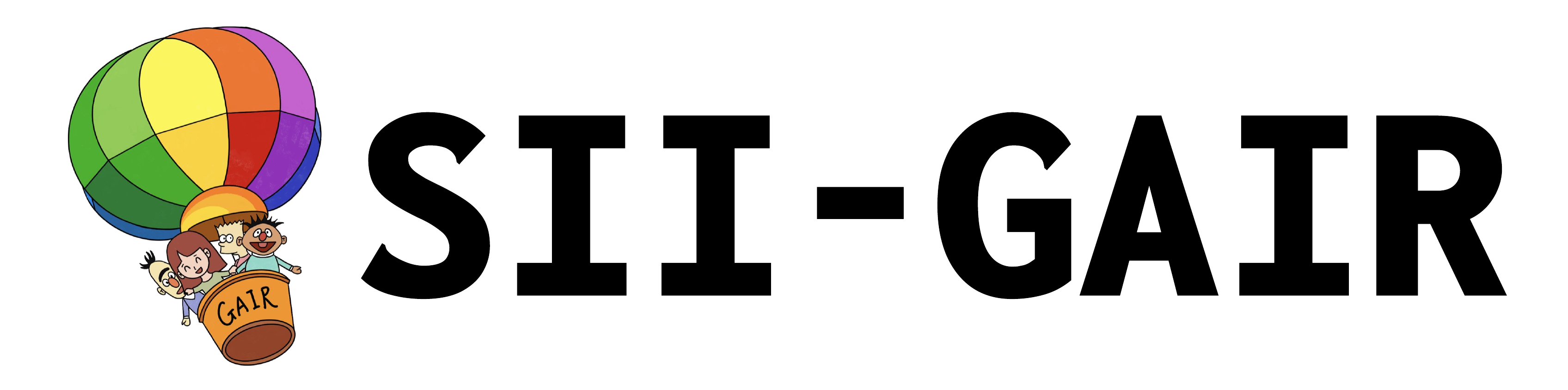}}%
}
\renewcommand{\headrulewidth}{0pt}
\setlength{\headsep}{2mm}

\renewcommand{\thefootnote}{}
\footnotetext{† Corresponding author.}
\vspace{-20pt}

{\centering
\quad \href{https://github.com/GAIR-NLP/DataEvolve}{\textcolor{black}\faGithub\ DataEvolve}
\quad \href{https://huggingface.co/datasets/GAIR/Darwin-CC}{\raisebox{-.15em}{\includegraphics[height=1em]{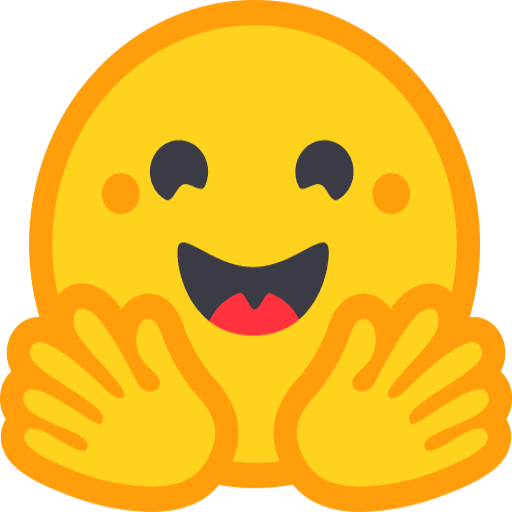}}\ Darwin-CC }
\par}
\vspace{10pt}


\begin{abstract}

Data Darwinism (Part I) established a systematic ten-level hierarchy for data processing, demonstrating that ascending this hierarchy—from basic filtering to model-driven enrichment—unlocks latent data value. However, that work relied on manually designed strategies for a single category. Modern pretraining corpora comprise hundreds of heterogeneous categories spanning domains, content types, and quality levels—each demanding specialized treatment. At this scale, manual strategy design becomes prohibitive: experts must analyze each category, identify quality issues, and iteratively refine approaches. This raises a fundamental question: \textbf{\textit{can strategies themselves evolve in an automated way?}}

We introduce \modelname, a framework that enables strategies to evolve through iterative optimization rather than manual design. For each data category, \modelname operates in a closed evolutionary loop: it observes data to identify quality issues, generates candidate strategies, executes them on sampled data, evaluates results with diagnostic feedback, and refines approaches across generations. This evolutionary process accumulates knowledge through an experience pool of discovered issues and a strategy pool tracking performance across iterations, enabling each generation to build upon previous insights.
Applied to 8 categories spanning 672B tokens from Nemotron-CC, \modelname produces \dataname, a 504B-token dataset with strategies evolved through 30 iterations per category. Training 3B models on 500B tokens, \dataname substantially outperforms raw data (+3.96 points) and achieves 44.13 average score across 18 benchmarks—surpassing DCLM (42.42), Ultra-FineWeb (36.29), and FineWeb-Edu (36.52), with pronounced gains on knowledge-intensive tasks (MMLU: +18.64, MedQA: +13.48).
Analysis reveals evolved strategies converge on cleaning-focused approaches: targeted noise removal and format normalization with domain-aware preservation—echoing the L4 (Generative Refinement) principles from Part I. Ablation studies confirm iterative evolution is essential: optimized strategies outperform suboptimal ones by 2.93 points, demonstrating that not all automated processing is equally effective. Our work establishes evolutionary strategy design as both feasible and necessary at pretraining scale, shifting the paradigm from manual expertise to automated evolution in data curation.

\end{abstract}

\begin{figure}[h]
    \centering
    \includegraphics[width=0.80\linewidth]{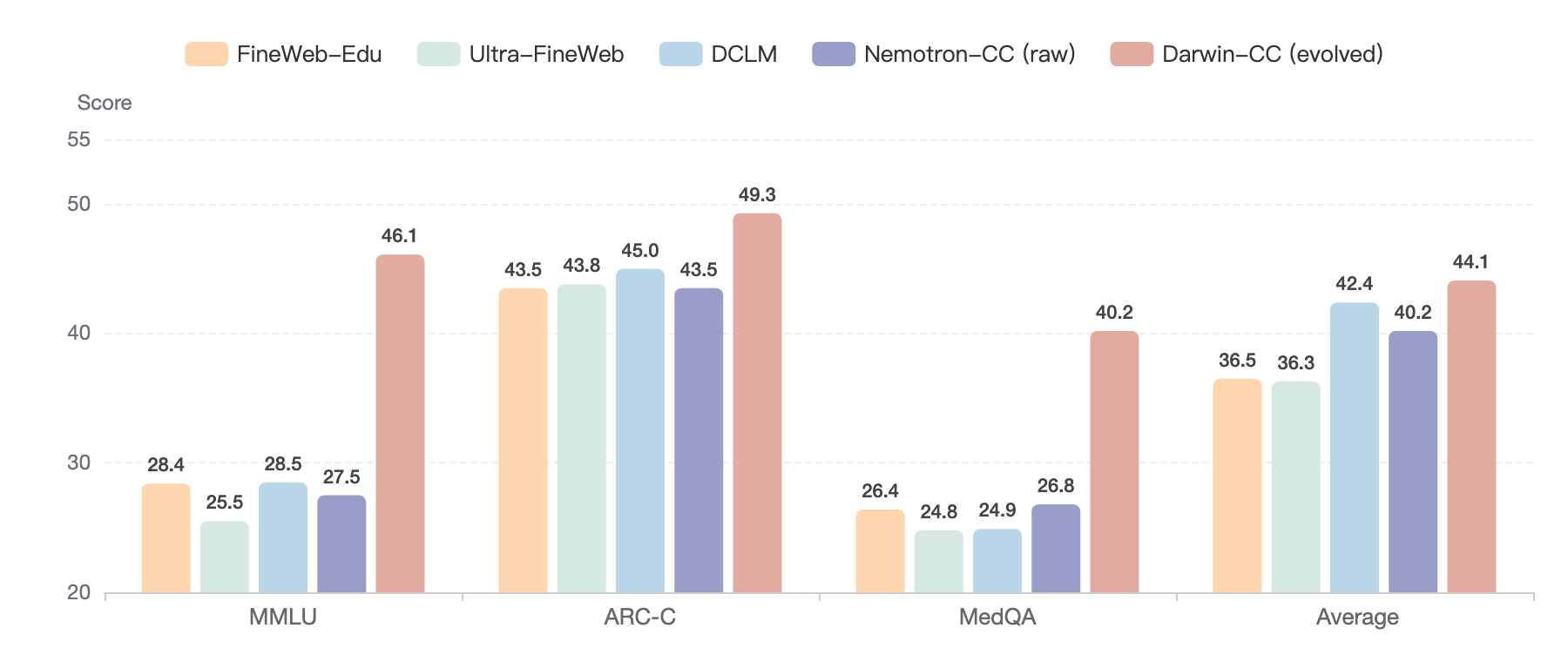}
    \caption{Performance comparison across pretraining datasets on 18 selected benchmarks.}
    \label{fig:teaser}
\end{figure}

\newpage
\pagestyle{fancy}
\renewcommand{\headrulewidth}{0.7pt}
\setlength{\headsep}{8mm}
\clearpage

\newpage

\renewcommand{\thefootnote}{\arabic{footnote}}
\setcounter{footnote}{0}

\section{Introduction}
\label{sec:introduction}



The performance of large language models is fundamentally determined by their pretraining data~\citep{longpre2023pretrainersguidetrainingdata}. Data Darwinism (Part I) ~\citep{qin2026datadarwinismiunlocking}established this principle through a systematic ten-level processing hierarchy (L0--L9), demonstrating that data quality is not static but evolves through progressively sophisticated transformations—from basic filtering (L0--L3) to model-driven refinement (L4--L5) and beyond. 
Part I validated this hierarchy on scientific literature, showing that raw data suffers from severe learnability gaps that only higher-level processing can bridge.
Yet the strategies driving this data evolution remained manually designed for a single category, leaving a critical question unanswered: \textit{can the strategies themselves evolve?}

Modern pretraining relies on web crawling to achieve trillion-token scale, but this introduces substantial noise. Meanwhile, high-quality curated sources such as textbooks and publications remain limited and grow slowly~\citep{villalobos2024rundatalimitsllm}. Quality-based filtering methods like FineWeb-Edu~\citep{penedo2024fineweb} and DCLM~\citep{li2024datacomplm} address this by retaining high-quality web pages while discarding low-quality content—an approach corresponding to the lower levels (L0--L3) of the Data Darwinism hierarchy. However, as Part I demonstrated, filtering-only approaches fail to unlock the full value of conceptually dense content. Aggressive filtering substantially reduces data quantity and risks losing valuable content obscured by surface-level noise, motivating data refinement: transforming noisy sources into usable material through higher-level processing~\citep{nguyen2025recyclingwebmethodenhance}.

Recent work demonstrates that large language models can automate data refinement at the L4--L5 levels by executing specified strategies at scale. Approaches including rephrasing web text into structured formats~\citep{karimi2025nemotroncc, maini2024rephrasing} and enriching content through guided rewriting~\citep{jiang2025generativedatarefinementjust} establish LLM-based curation feasibility. Modern pretraining corpora span diverse categories with distinct characteristics, and recent work explores tailored strategies: SwallowCode and SwallowMath~\citep{fujii2025rewritingpretrainingdataboosts} for code and mathematics, MegaMath~\citep{zhou2025megamath} for mathematical web pages, and MegaScience~\citep{fan2025megasciencepushingfrontiersposttraining} with discipline-specific extraction strategies.

While these tailored approaches demonstrate clear benefits, they reveal a fundamental scalability bottleneck. Designing effective strategies requires analyzing category-specific characteristics, identifying quality issues, formulating appropriate operations, and iteratively refining approaches—demanding substantial expertise per category. 
Even adding a single few-shot example to a curation strategy can be painstaking, as it may require sifting through dozens of candidate examples before identifying a single high-quality one that effectively guides the model ~\citep{Jiang2022PromptMakerPP}. Moreover, even validating a single candidate strategy requires cleaning the data and training a model at scale, making iterative refinement practically infeasible ~\citep{li2024datacomplm}. However, modern pretraining corpora comprise hundreds or thousands of categories formed by intersecting subject domains, content types, quality levels, and other dimensions, each exhibiting distinct characteristics~\citep{ai2025essentialwebv1024ttokens, wettig2025organizewebconstructingdomains}. At this scale, comprehensive per-category manual design becomes prohibitive. This raises our central question: \textit{can strategies themselves evolve in an
automated way?} Recent advances show AI systems can autonomously design solutions—from optimizing prompts~\citep{zhou2023largelanguagemodelshumanlevel} to discovering architectures~\citep{liu2025alphagomomentmodelarchitecture}—suggesting that strategy design, like the data it processes, can undergo evolutionary refinement.

We introduce \modelname, a framework enabling strategies to evolve through iterative optimization rather than manual design. \modelname operates in a closed evolutionary loop: for each category, it observes data to identify quality issues, generates candidate strategies, executes them on sampled documents, and refines approaches based on diagnostic feedback. Rather than full model training, \modelname approximates strategy fitness through sample-based quality assessment, enabling rapid iterative refinement without prohibitive training overhead. Critically, the system evolves knowledge across generations through two mechanisms: an \textit{experience pool} accumulating discovered quality issues, and a \textit{strategy pool} tracking performance and analyses across iterations. Each generation builds upon previous insights, with successful strategies serving as parents for the next generation while unsuccessful ones are pruned—a natural selection process for data curation strategies. This transforms strategy design from a manual, expertise-intensive process into an evolutionary optimization that scales across hundreds of categories.

We apply \modelname to 8 categories spanning 672B tokens from Nemotron-CC~\citep{karimi2025nemotroncc}, evolving strategies through 30 generations per category to produce \dataname, a 504B-token dataset. Training 3B models on 500B tokens, \dataname substantially outperforms raw data (+3.96 points average across 18 benchmarks) and gains are particularly pronounced on knowledge-intensive tasks (MMLU +18.64, MedQA +13.48), and achieves 44.13 average score, surpassing established corpora including DCLM (42.42), Ultra-FineWeb (36.29), and FineWeb-Edu (36.52) under identical training budgets.

Ablation studies confirm that evolution is essential: optimized strategies outperform suboptimal ones by 2.93 points, demonstrating that not all automated processing is equally effective—only evolved strategies unlock full data value, and that our sample-based fitness approximation reliably reflects true downstream performance. Analysis reveals that evolved strategies converge on \textit{cleaning-focused approaches}: targeted noise removal with domain-aware preservation, echoing the L4 (Generative Refinement) principles from Part I. This convergence demonstrates that systematic cleaning, when properly evolved for each category's characteristics, suffices for substantial quality improvements without expensive content transformation.

\textbf{Our contributions are:} 
\textbf{(1)} We demonstrate that strategies can evolve: by formulating strategy design as an evolutionary optimization problem with diagnostic feedback and cross-generation knowledge accumulation, we enable automated discovery of effective curation strategies validated at pretraining scale.
\textbf{(2)} We release \modelname, an end-to-end framework for evolutionary strategy design, and \dataname, a 504B-token dataset where each category's strategy has evolved through 30 generations, outperforming established corpora.
\textbf{(3)} We reveal evolutionary convergence: independently evolved strategies across diverse categories converge on cleaning-focused approaches with domain-aware preservation, offering a simpler and more scalable path than transformation-based methods for pretraining data curation.

\section{Problem Formulation}
\label{sec:problem}

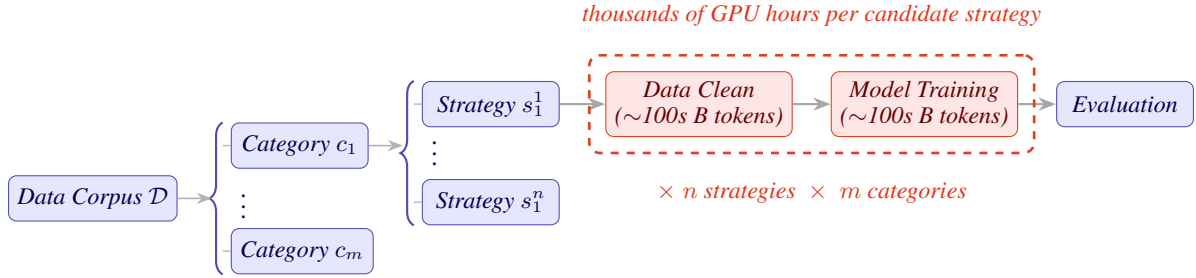
\begin{figure}[t]
\centering
\begin{tikzpicture}[
    node distance=0.3cm and 0.5cm,
    box/.style={
        rectangle,
        rounded corners=3pt,
        draw=blue!40!gray,
        fill=blue!10,
        text=blue!30!black,
        minimum width=1.8cm,
        minimum height=0.6cm,
        align=center,
        font=\small\itshape
    },
    expensive/.style={
        rectangle,
        rounded corners=3pt,
        draw=red!50!brown,
        fill=red!10,
        text=red!40!black,
        minimum width=1.8cm,
        minimum height=0.6cm,
        align=center,
        font=\small\itshape
    },
    arrow/.style={
        ->,
        >=Stealth,
        thick,
        gray!70
    },
    brace arrow/.style={
        ->,
        >=Stealth,
        semithick,
        gray!60
    },
    annotation/.style={
        font=\footnotesize\itshape,
        text=red!60!brown,
        align=center
    }
]

\node[box] (corpus) {Data Corpus $\mathcal{D}$};

\node[box, right=0.7cm of corpus, yshift= 0.7cm] (cat1) {Category $c_1$};
\node[box, right=0.7cm of corpus, yshift=-0.7cm] (catm) {Category $c_m$};
\node[text=blue!40!black, font=\normalsize,
      right=0.7cm of corpus, yshift=0.0cm] (catdots) {$\vdots$};

\draw[decorate, decoration={brace, amplitude=5pt, mirror},
      blue!40!gray, thick]
    ([xshift=-3pt]cat1.north west) -- ([xshift=-3pt]catm.south west)
    node[midway, left=8pt, text=blue!30!black, font=\footnotesize\itshape] {};

\draw[brace arrow] (corpus.east) -- ++(0.45,0);

\draw[gray!50, semithick] ([xshift=-3pt]cat1.west) -- (cat1.west);
\draw[gray!50, semithick] ([xshift=-3pt]catm.west) -- (catm.west);

\node[box, right=0.7cm of cat1, yshift= 0.55cm] (s1) {Strategy $s_1^1$};
\node[box, right=0.7cm of cat1, yshift=-0.75cm] (sn) {Strategy $s_1^n$};
\node[text=blue!40!black, font=\normalsize,
      right=0.7cm of cat1, yshift=0.0cm] (sdots) {$\vdots$};

\draw[decorate, decoration={brace, amplitude=5pt, mirror},
      blue!40!gray, thick]
    ([xshift=-3pt]s1.north west) -- ([xshift=-3pt]sn.south west);

\draw[brace arrow] (cat1.east) -- ++(0.45,0);

\draw[gray!50, semithick] ([xshift=-3pt]s1.west) -- (s1.west);
\draw[gray!50, semithick] ([xshift=-3pt]sn.west) -- (sn.west);

\node[expensive, right=0.6cm of s1] (clean)
    {Data Clean\\{\footnotesize($\sim$100s B tokens)}};
\draw[arrow] (s1.east) -- (clean.west);

\node[expensive, right=0.5cm of clean] (train)
    {Model Training\\{\footnotesize($\sim$100s B tokens)}};
\draw[arrow] (clean.east) -- (train.west);

\node[box, right=0.5cm of train] (eval) {Evaluation};
\draw[arrow] (train.east) -- (eval.west);

\node[
    draw=red!50!brown,
    dashed,
    rounded corners,
    line width=1.0pt,
    fit=(clean)(train),
    inner sep=6pt,
] (expensivebox) {};

\node[annotation, above=0.25cm of expensivebox]
    {thousands of GPU hours per candidate strategy};
\node[annotation, below=0.25cm of expensivebox]
    {$\times\; n \text{ strategies} \;\times\; m \text{ categories}$};

\end{tikzpicture}
\caption{Naively evaluating candidate curation strategies requires
cleaning data at full scale and training a model to convergence for
each candidate strategy—demanding thousands of GPU hours per evaluation.
Across $n$ candidate strategies and $m$ categories, the total cost
becomes computationally intractable.}
\label{fig:eval-pipeline}
\end{figure}

Modern pretraining corpora span hundreds of heterogeneous categories, each with distinct quality characteristics requiring tailored curation strategies. We formalize the problem of designing effective strategies at this scale.

Let $\mathcal{C} = \{c_1, c_2, \ldots, c_m\}$ denote a set of data categories, where each category $c_i$ contains a corpus $\mathcal{D}_i = \{d_1, d_2, \ldots, d_n\}$ of raw documents. A curation strategy $s_i$ for category $c_i$ specifies processing operations—what to remove, preserve, and normalize—that an LLM-based executor $f(\cdot, \cdot)$ applies to produce cleaned data:
\begin{equation}
\mathcal{D}'_i = \{f(d, s_i) \mid d \in \mathcal{D}_i\}
\end{equation}

Ideally, strategy quality would be determined by downstream model performance. Let $M_{\mathcal{D}'}$ denote a language model trained on $\mathcal{D}' = \bigcup_{i=1}^{m} \mathcal{D}'_i$ with performance $Q(M)$ on evaluation benchmarks. The optimal strategies would satisfy:
\begin{equation}
(s^*_1, \ldots, s^*_m) = \argmax_{s_1, \ldots, s_m} Q(M_{\mathcal{D}'})
\end{equation}

However, directly optimizing this objective is prohibitively \textbf{expensive} (Figure \ref{fig:eval-pipeline}): each evaluation requires cleaning data, training a model from scratch, and measuring performance—demanding thousands of GPU hours. Moreover, quality differences between strategies often emerge only at scales of hundreds of billions of tokens during training. This cost is further compounded in the pretraining setting: unlike fine-tuning where quality differences emerge quickly, pretraining requires processing hundreds of billions of tokens before the effect of a curation strategy becomes statistically distinguishable, rendering exhaustive search over candidate strategies computationally intractable.

To make the problem tractable, we approximate the objective using a quality function $S(\mathcal{D}'_i)$ that evaluates cleaned data through sample-based assessment (rating document quality on a 1-10 scale):
\begin{equation}
s^*_i = \argmax_{s_i} S(\mathcal{D}'_i)
\end{equation}

This reformulation enables iterative strategy refinement without model training at each step. However, the problem remains challenging: the strategy space $\mathcal{S}$ is discrete and combinatorially large, with strategies varying in operation choices, decision criteria, and quality issue coverage. Moreover, effective design requires analyzing category-specific characteristics, identifying quality issues, and iterative refinement—effort that compounds across hundreds of heterogeneous categories.

\section{\modelname: Evolutionary Strategy Design}
\label{sec:framework}
\subsection{Overview}
\begin{figure}[t]
    \centering
    \includegraphics[width=0.99\linewidth]{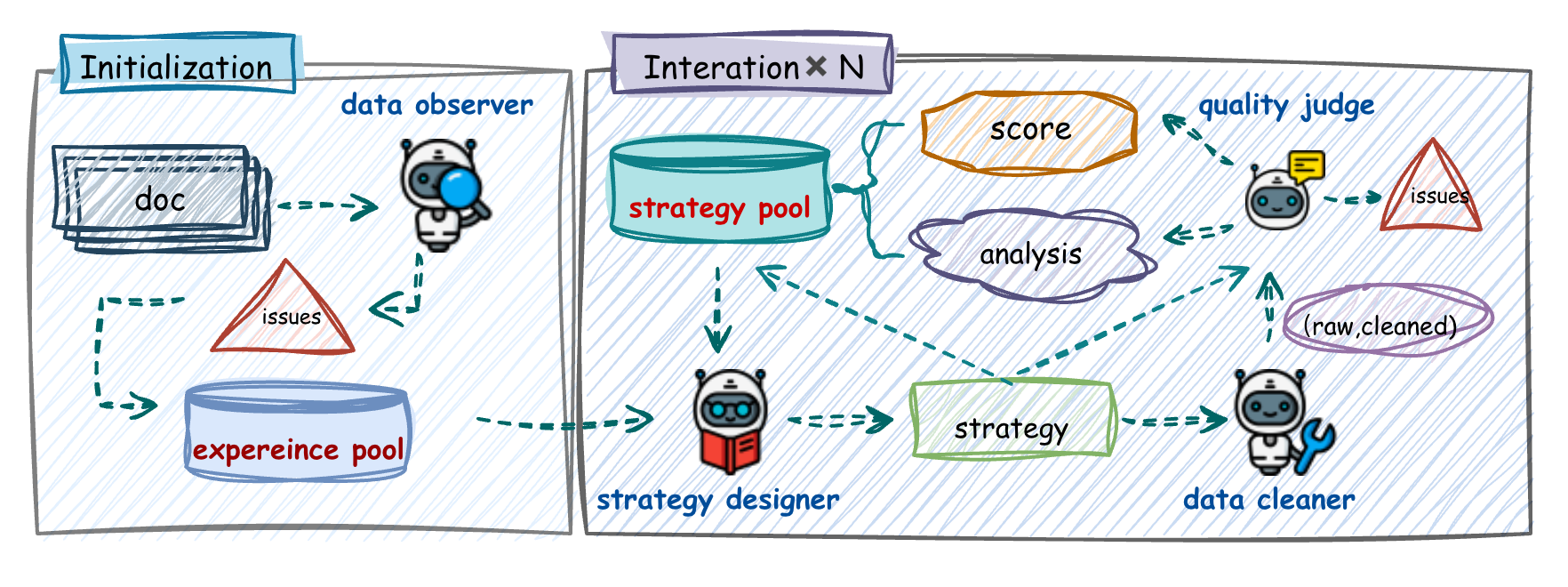}
 \caption{Overview of the \modelname~framework. The system enables strategies to evolve through an iterative feedback loop involving four core components: (1) the \textit{data observer} identifies category-specific quality issues; (2) the \textit{strategy designer} generates and refines cleaning strategies; (3) the \textit{data cleaner} executes strategies on sample data; and (4) the \textit{quality judge} provides scoring and diagnostic feedback. Discovered issues and evolved strategies are archived in the \textit{experience pool} and \textit{strategy pool}, enabling cross-generation knowledge transfer to guide evolutionary progression.}
    \label{fig:pipeline}
\end{figure}

\modelname~enables category-specific strategies to evolve through iterative optimization rather than manual design. The framework operates as an evolutionary system with four core components in a closed feedback loop: a \textit{data observer} identifies quality issues in sampled data, a \textit{strategy designer} generates and refines curation strategies, a \textit{data cleaner} executes these strategies, and a \textit{quality judge} evaluates results and provides diagnostic feedback. Two repositories enable cross-generation knowledge transfer: an \textit{experience pool} accumulates discovered quality issues across all iterations, and a \textit{strategy pool} tracks evolved strategies with their performance scores and diagnostic analyses.

The evolutionary process operates in two phases. During \textit{initialization}, the system observes the target category to identify quality problems, establishing the initial knowledge base. Through \textit{evolutionary refinement}, strategies evolve over multiple generations: each iteration selects the best-performing strategy as a parent, generates refined variants through mutation (guided by diagnostic feedback), evaluates offspring on new samples, and propagates successful strategies to the next generation. Each category evolves independently, enabling specialized strategies tailored to distinct data characteristics. The complete evolutionary workflow is illustrated in Figure~\ref{fig:pipeline}.




\subsection{Initialization: Establishing the Knowledge Base}

Before strategies can evolve, \modelname~establishes an initial understanding of the target category. The \textit{data observer} examines sampled documents and produces a structured assessment revealing category-specific quality patterns—such as equation rendering errors and broken citations in stem content, ambiguous medical terminology in medicine, or code snippets mixed with corrupted symbols in computer science.

These initial observations seed the \textit{experience pool}—a repository of quality issues that guides all subsequent evolutionary cycles. Rather than each generation starting from scratch, this accumulated knowledge ensures that evolved strategies address both initially observed problems and issues discovered during execution, enabling genuine evolutionary progress rather than random search.

\subsection{Evolutionary Strategy Optimization}

With the knowledge base established, \modelname~enters its core evolutionary loop: generate strategy variants, execute them on sample data, evaluate fitness, and select survivors for the next generation. This cycle repeats for multiple generations, with each generation building upon the insights of its predecessors.


\paragraph{Strategy generation.} The \textit{strategy designer} operates differently across iterations. In the first iteration, it synthesizes observations from the \textit{experience pool} into an initial strategy. From the second iteration onward, it refines the best-performing strategy from the \textit{strategy pool}, guided by diagnostic feedback explaining what worked, what gaps remain, and what needs clarification.


\paragraph{Execution.} The \textit{data cleaner} applies the designed strategy to a batch of sampled documents. To ensure generalization, different documents are sampled in each iteration, preventing strategies from overfitting to specific examples. This produces pairs of original and cleaned documents that reveal the strategy's actual behavior.


\paragraph{Evaluation.} The \textit{quality judge} evaluates results in three steps. First, it scores each (original, cleaned) pair on a 1-10 scale with comments on improvements and remaining issues; the average becomes the strategy's overall score. Second, it analyzes strategy coverage (which issues from the \textit{experience pool} were addressed) and executability (whether instructions were clear and consistently followed). Third, it often discovers new quality issues during evaluation, which are added to the \textit{experience pool} for future iterations.

\paragraph{Closing the loop.} After each iteration, results are stored in both repositories: the \textit{strategy pool} records the strategy with its score and diagnostic analysis, while the \textit{experience pool} accumulates newly discovered issues. When the next iteration begins, the \textit{strategy designer} accesses both an enriched \textit{experience pool} (with more quality issues identified) and the best-performing strategy with its diagnostic feedback from the strategy pool. This feedback loop enables each iteration to produce more targeted and effective strategies than the previous one. After completing all iterations, the highest-scoring strategy from the \textit{strategy pool} is selected for deployment on the full dataset.

\section{Experiments}

\subsection{Dataset Construction}

\paragraph{Data Source} We construct our dataset from Nemotron-CC~\citep{karimi2025nemotroncc}, a large-scale web corpus derived from Common Crawl. We use only the real (non-synthetic) part, which provides quality annotations at five levels. To simplify our experimental design, we merge these into two categories: \textit{high} and \textit{not-high}.

Beyond quality, we require finer-grained categorization to enable domain-specific prompt design. We employ the \texttt{EAI-Distill-0.5B} ~\citep{ai2025essentialwebv1024ttokens}, a specialized document classification model fine-tuned from Qwen2.5-0.5B-Instruct that provides annotations across 12 taxonomic dimensions, including Free Decimal Correspondence (FDC), Bloom's Taxonomy, Document Type, Content Quality, and Educational Metadata. We focus on two of these dimensions for our categorization scheme.(1) For the \textit{FDC }, which captures subject domain, we perform a mapping to derive five broad disciplines: \textit{human-social}, \textit{mathematics}, \textit{computer science}, \textit{medicine}, and \textit{other stem}. (2) Similarly, for the \textit{Document Type}, we map the original classifications into five content categories: \textit{academic}, \textit{code}, \textit{fragment}, \textit{social media}, and \textit{text}. Details of these mappings are provided in Appendix~\ref{app:fdc-mapping}.

For our experiments, we select a focused subset from this categorization space: we use only \textit{academic} content across four technical disciplines (\textit{mathematics}, \textit{computer science}, \textit{medicine}, and \textit{other stem}), each at two quality levels (\textit{high}, \textit{not-high}). This selection yields 8 distinct categories comprising approximately 672B tokens in total.

\begin{wraptable}{r}{0.5\textwidth}  
\centering
\small
\setlength{\tabcolsep}{3pt}
\caption{Category composition of \dataname}
\label{tab:category_distribution}
\begin{tabular}{lllrr}
\toprule
\textbf{Content Type} & \textbf{Quality} & \textbf{Domain} & \textbf{Raw} & \textbf{Cleaned} \\
\midrule
\multirow{8}{*}{Academic} 
    & \multirow{4}{*}{High} 
        & Mathematics      & 7B  & 5B \\
    &   & Computer Science & 32B  & 24B \\
    &   & Medicine         & 52B  & 29B \\
    &   & Other STEM       & 71B  & 51B \\
\cmidrule{2-5}
& \multirow{4}{*}{Not-High}
        & Mathematics      & 6B & 6B \\
    &   & Computer Science & 107B  & 78B \\
    &   & Medicine         & 164B  & 132B \\
    &   & Other STEM       & 233B  & 179B \\
\midrule
\multicolumn{3}{l}{\textbf{Total}} & \textbf{672B} & \textbf{504B} \\
\bottomrule
\end{tabular}
\end{wraptable}

\paragraph{\modelname Configuration}
We apply \modelname independently to each of the 8 categories for 30 iterations. Each iteration operates as follows: the \textit{data observer}, implemented with \texttt{GPT-4o-mini}~\citep{GPT-4omini}, analyzes 100 sampled documents to identify quality issues; the \textit{strategy designer}, using \texttt{o4-mini}~\citep{o4mini}, synthesizes these observations with accumulated feedback to generate a refined cleaning prompt; the \textit{data cleaner} executes this prompt via \texttt{gpt-oss-120b}~\citep{openai2025gptoss120bgptoss20bmodel} on 500 documents; and the \textit{quality judge}, also using \texttt{gpt-5-mini}~\citep{GPT-5}, evaluates 50 randomly sampled pairs from the cleaned results, producing both scores and diagnostic feedback.

\paragraph{Final Dataset.}
After 30 iterations per category, we select highest-scoring strategy from each strategy pool for deployment. Each optimized strategy is then applied to its corresponding full category corpus using the same LLM-based data cleaner. This large-scale cleaning process, executed on the complete 672B-token raw data, produces \dataname, comprising 504B tokens. The 25\% token reduction reflects targeted removal of low-quality content—duplicates, formatting artifacts, incomplete fragments, and noise—while preserving domain-specific valuable content through the discovered preservation rules. Table~\ref{tab:category_distribution} shows the token distribution across the 8 categories before and after cleaning. We do not perform explicit benchmark decontamination, consistent with the practice of the upstream Nemotron-CC corpus and comparable datasets~\citep{karimi2025nemotroncc,li2024datacomplm,lozhkov2024fineweb-edu}. We note that our cleaning process targets removing artifacts, duplicates, and 
low-quality fragments, and thus is unlikely to systematically alter benchmark contamination levels relative to the source corpus.

\subsection{Training Configuration}

To ensure fair comparison, we train all models from scratch on 500B tokens using identical configurations. We employ a 3B-parameter Qwen2.5 architecture~\citep{qwen2.5} with 4,096-token context length. Training uses AdamW~\citep{loshchilov2019decoupled} with a learning rate that warms up to $3 \times 10^{-4}$ over 6,000 steps and remains constant thereafter, a global batch size of 1,024, and runs for 120,000 steps total, approximately 500B tokens.

\subsection{Evaluation Benchmarks}

We evaluate all trained models on a comprehensive benchmark suite, including MMLU (5-shot)~\citep{hendrycks2020measuring}, ARC-Easy and ARC-Challenge (10-shot)~\citep{clark2018think}, OpenBookQA (5-shot)~\citep{OpenBookQA2018}, PIQA (10-shot)~\citep{bisk2020piqa}, HellaSwag (10-shot)~\citep{zellers2019hellaswag}, WinoGrande (0-shot)~\citep{ai2:winogrande}, SIQA (5-shot)~\citep{sap2019socialIQa}, RACE (0-shot)~\citep{lai2017racelargescalereadingcomprehension}, BBH (3-shot)~\citep{suzgun2022challenging}, DROP (5-shot)~\citep{dua2019drop}, AGIEval-En (0-shot)~\citep{zhong2023agieval}, CSQA(0-shot)~\citep{talmor-etal-2019-commonsenseqa}, TriviaQA (5-shot)~\citep{2017arXivtriviaqa}, GPQA-Main (5-shot)~\citep{rein2024gpqa}, MedQA (0-shot)~\citep{jin2021disease}, MedMCQA (0-shot)~\citep{pal2022medmcqa}, and PubMedQA (0-shot)~\citep{jin2019pubmedqa}, totally 18 benchmarks. All evaluations are conducted using the \texttt{lm-evaluation-harness} ~\citep{eval-harness} framework.

\section{Results}
\subsection{Validating Automated Strategy Design}

\begin{table}[htbp]
\centering
\setlength{\tabcolsep}{12pt} 
\caption{Performance comparison across curation strategies. Models trained on raw data (Raw), data with suboptimal strategy (Sub-opt), data cleaned with \modelname's best strategy (Best), which generates \dataname. All models are 3B parameters trained on 500B tokens.}
\label{tab:benchmark_comparison_prompt}
\begin{tabular}{lccc}
\toprule
\textbf{Metrics} & \textbf{Raw} & \textbf{Sub-opt} & \textbf{Best} \\
\midrule
BBH            & 26.82 & 26.69\textsubscript{\color{deepgreen}-0.12}  & 26.16\textsubscript{\color{deepgreen}-0.65} \\
ARC-E           & 74.94 & 77.55\textsubscript{\color{red}+2.61} & 78.59\textsubscript{\color{red}+3.65} \\
ARC-C           & 43.52 & 48.41\textsubscript{\color{red}+4.90} & 49.32\textsubscript{\color{red}+5.80} \\
MMLU   & 27.49 & 32.55\textsubscript{\color{red}+5.05}  & 46.13\textsubscript{\color{red}+18.64} \\
AGIEval       & 18.15 & 18.30\textsubscript{\color{red}+0.15}  & 18.21\textsubscript{\color{red}+0.07} \\
HellaSwag      & 65.32 & 64.36\textsubscript{\color{deepgreen}-0.96}  & 62.21\textsubscript{\color{deepgreen}-3.11} \\
TriviaQA            & 25.33 & 26.96\textsubscript{\color{red}+1.63}  & 26.65\textsubscript{\color{red}+1.32} \\
RACE            & 35.04 & 35.63\textsubscript{\color{red}+0.59}  & 34.28\textsubscript{\color{deepgreen}-0.76} \\
DROP   & 19.57 & 19.48\textsubscript{\color{deepgreen}-0.09}  & 18.49\textsubscript{\color{deepgreen}-1.08} \\
WinoGrande       & 57.96 & 59.89\textsubscript{\color{red}+1.93}  & 58.09\textsubscript{\color{red}+0.13} \\
PIQA      & 76.79 & 76.80\textsubscript{\color{red}+0.01}  & 76.15\textsubscript{\color{deepgreen}-0.64} \\
CSQA            & 20.31 & 20.61\textsubscript{\color{red}-0.30}  & 39.12\textsubscript{\color{red}+18.80} \\
SIQA            & 44.36 & 43.58\textsubscript{\color{deepgreen}-0.78}  & 43.57\textsubscript{\color{deepgreen}-0.79} \\
OpenBookQA       & 39.80 & 42.20\textsubscript{\color{red}+2.40}  & 41.44\textsubscript{\color{red}+1.64} \\
GPQA-Main      & 24.37 & 23.93\textsubscript{\color{deepgreen}-0.45}  & 27.10\textsubscript{\color{red}+2.72} \\
MedQA            & 26.77 & 26.11\textsubscript{\color{deepgreen}-0.66}  & 40.25\textsubscript{\color{red}+13.48} \\
MedMCQA            & 28.86 & 30.28\textsubscript{\color{red}+1.42}  & 40.97\textsubscript{\color{red}+12.10} \\
PubMedQA      & 67.68 & 68.32\textsubscript{\color{red}+0.64}  & 67.68\textsubscript{\color{red}+0.00} \\
\midrule
\textit{Average} & 40.17 & 41.20\textsubscript{\color{red}+1.03} & 44.13\textsubscript{\color{red}+3.96} \\
\bottomrule
\end{tabular}
\end{table}

\begin{figure}[htbp]
\centering
\begin{minipage}[t]{0.48\columnwidth}
    \centering
    \includegraphics[width=\textwidth]{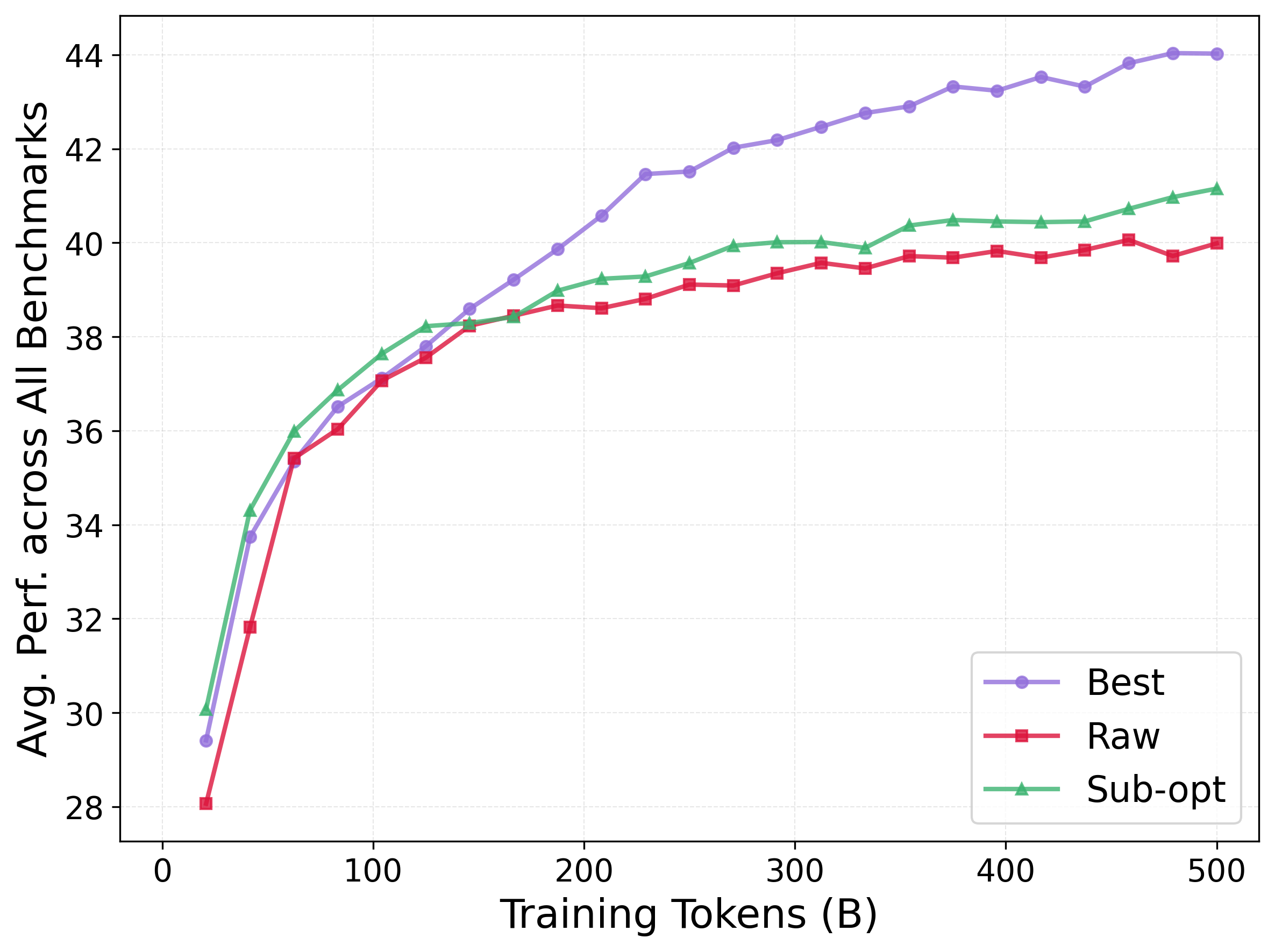}
    \caption{Learning curves of 3B models trained on raw data, suboptimal strategy, and optimized strategy (\dataname) over 500B tokens.}
    \label{fig:training_curves-val}
\end{minipage}
\hfill
\begin{minipage}[t]{0.48\columnwidth}
    \centering
    \includegraphics[width=\textwidth]{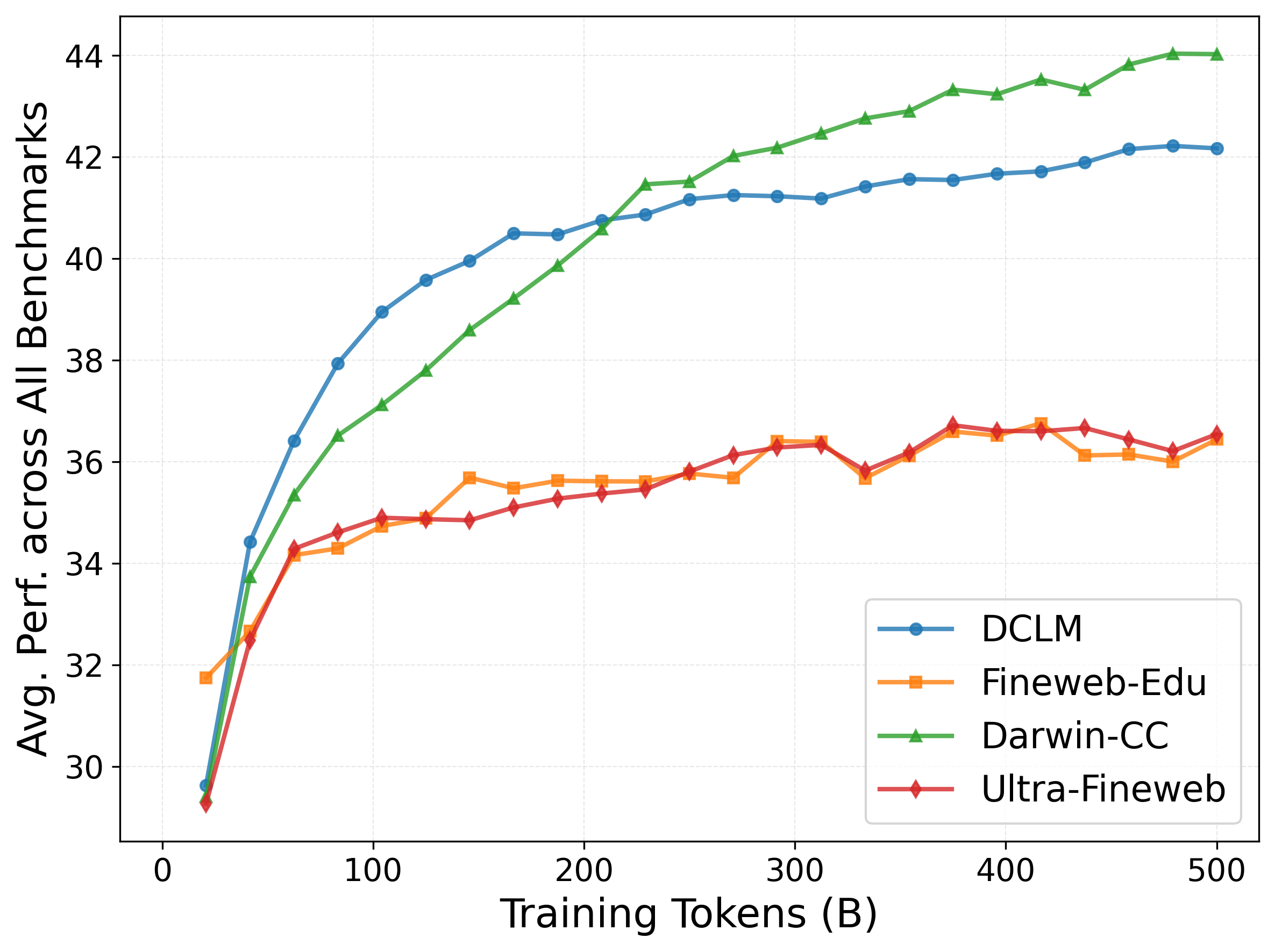}
    \caption{Learning curves comparison across pretraining corpora. All models are 3B parameters trained for 500B tokens.}
    \label{fig:performance_diff_dataset}
\end{minipage}
\end{figure}

To validate \modelname's effectiveness, we compare three data configurations under identical training setups (3B model, 500B tokens): raw unprocessed data, data cleaned with a lower-scoring strategy from the \textit{strategy pool}, and \dataname cleaned with \modelname's best-performing strategies. Table~\ref{tab:benchmark_comparison_prompt} presents results across 18 benchmarks, and Figure~\ref{fig:training_curves-val} shows the learning curves throughout training.

\paragraph{Automated design improves data quality.} Models trained on \dataname achieve an average score of 44.13, substantially outperforming raw data by 3.96 points. This consistent improvement demonstrates that automated strategy design effectively enhances pretraining data quality. Gains are particularly pronounced on knowledge-intensive tasks: MMLU improves by 18.64 points, CSQA by 18.80 points, and MedMCQA by 12.11 points, suggesting that \modelname successfully preserves factual information while removing noise. Beyond final performance, Figure~\ref{fig:training_curves-val} reveals superior data efficiency throughout training. \dataname achieves steeper early improvements and maintains widening advantages as training progresses—the gap over raw data grows from 2.49 points at 250B tokens to 3.96 points at 500B tokens. This scaling behavior demonstrates that data quality differences compound during training, with optimized strategies enabling more effective learning at scale.

\paragraph{Iterative optimization is essential.} Comparison with the lower-scoring strategy reveals that not all automated cleaning is equally effective. Data cleaned with the suboptimal strategy achieves only 41.20 average score, barely above raw data (+1.03 points) and significantly below \dataname(gap: 2.93 points). This pattern is also especially stark on knowledge tasks: on MMLU, \dataname gains +18.64 over raw data while the suboptimal strategy gains only +5.05, leaving a 13.59-point gap attributable to strategy quality alone. These disparities confirm that iterative optimization, not just LLM-based processing, drives effective data curation.

\subsection{Comparison with Established Pretraining Corpora}

To assess the practical value of \dataname, we compare it against three widely-used pretraining datasets: FineWeb-Edu~\citep{lozhkov2024fineweb-edu}, a high-quality educational content subset; Ultra-FineWeb~\citep{wang2025ultrafineweb}, an aggressively filtered version emphasizing quality over scale; and DCLM~\citep{li2024datacomplm}, a large-scale curated corpus using classifier-based filtering. All models are trained under identical configurations (3B parameters, 500B tokens) and evaluated on 18 benchmarks. Table~\ref{tab:benchmark_comparison_across_different_dataset} presents the results.

\dataname achieves the highest average score at 44.13, outperforming the second-best baseline DCLM by 1.71 points. This improvement is particularly pronounced on knowledge-intensive tasks such as MMLU. Performance is generally competitive or superior across most benchmarks, though a few tasks such as HellaSwag and TriviaQA show variability reflecting different curation emphases. Overall, these results demonstrate that \dataname achieves quality competitive with or superior to established pretraining datasets, validating both \modelname's automated strategy design approach and the practical value of the resulting corpus.

Beyond final performance, Figure~\ref{fig:performance_diff_dataset} show that \dataname achieves steeper improvements in early training stages and maintains consistent gains as training progresses, suggesting that automated strategy design produces not only higher-quality data but also more sample-efficient pretraining corpora.

\begin{table}[htbp]
\centering
\setlength{\tabcolsep}{3pt}
\caption{Comparison of models trained on different pretraining corpora. All models are 3B parameters trained on 500B tokens.}
\label{tab:benchmark_comparison_across_different_dataset}
\begin{tabular}{lcccc}
\toprule
\textbf{Metrics} & \textbf{Fineweb-Edu} & \textbf{Ultra-Fineweb} & \textbf{DCLM} & \textbf{\dataname} \\
\midrule
BBH             & 3.01  & 7.42  & 24.16 & \textbf{26.16} \\
ARC-E           & 73.39 & 73.96 & 75.13 & \textbf{78.59} \\
ARC-C           & 43.45 & 43.77 & 45.02 & \textbf{49.32} \\
MMLU            & 28.38 & 25.53 & 28.54 & \textbf{46.13} \\
AGIEval         & 16.96 & 17.72 & 17.90 & \textbf{18.21} \\
HellaSwag       & 64.33 & 65.32 & \textbf{70.39} & 62.21 \\
TriviaQA        & 0.67  & 0.42  & \textbf{42.85} & 26.65 \\
RACE            & 35.43 & 34.28 & \textbf{36.08} & 34.28 \\
DROP            & 6.78  & 7.78  & \textbf{24.31} & 18.49 \\
WinoGrande      & 61.02 & 60.93 & \textbf{64.99} & 58.09 \\
PIQA            & 75.80 & 75.63 & \textbf{77.93} & 76.15 \\
CSQA            & 19.54 & 19.90 & 20.16 & \textbf{39.12} \\
SIQA            & 45.02 & 43.43 & \textbf{47.51} & 43.57 \\
OpenBookQA      & 39.92 & 39.84 & \textbf{43.36} & 41.44 \\
GPQA-Main       & 24.51 & 23.04 & 25.67 & \textbf{27.10} \\
MedQA           & 26.36 & 24.84 & 24.88 & \textbf{40.25} \\
MedMCQA         & 25.80 & 24.92 & 28.15 & \textbf{40.97} \\
PubMedQA        & 67.04 & 64.44 & 66.56 & \textbf{67.68} \\
\midrule
\textit{Average} & 36.52 & 36.29 & 42.42 & \textbf{44.13} \\
\bottomrule
\end{tabular}
\end{table}
\section{Analysis}

\subsection{What Do \modelname Strategies Actually Do?}
Examining the category-specific strategies reveals a clear pattern: \modelname learns to clean, not to transform. Contrary to prior approaches that rephrase web text into idealized formats (textbook-style exposition, Wikipedia articles, or question-answer pairs), discovered strategies focus on removing unwanted content and normalizing formatting while preserving original text.

The cleaning operations fall into three categories: (1) removing web artifacts (HTML tags, navigation menus, advertisements, duplicate sentences, PII); (2) normalizing formatting (whitespace, punctuation, numeric expressions); and (3) applying domain-specific preservation rules. These preservation rules reveal category-specific tailoring: \textit{stem} strategies preserve scientific nomenclature and equations with LaTeX handling; \textit{mathematics} strategies maintain theorem statements and proofs; \textit{medicine} strategies preserve clinical units and drug names; \textit{computer science} strategies protect code blocks and technical syntax.

This cleaning-focused approach yields substantial improvements by removing artifacts and normalizing formatting errors while preserving original content. The score gain demonstrates that systematic cleaning enhances quality without rewriting text into homogeneous formats such as textbook-style exposition or question-answer pairs. Unlike transformation-based approaches that impose uniform writing styles, cleaning preserves the diverse content and structure of web sources while removing technical corruption. This offers a practical path for scaling data curation to heterogeneous pretraining corpora.

\begin{wraptable}{r}{0.45\textwidth}  
\centering
\caption{Diversity metrics for raw and cleaned datasets.}
\label{tab:diversity_metrics}
\small
\begin{tabular}{lccr}
\toprule
\textbf{Metric} & \textbf{Raw} & \textbf{Cleaned} & \textbf{Change} \\
\midrule
Self-ROUGE-2 ↓       & 0.0148 & 0.0116 & \textbf{-21.7\%} \\
L2 Distance ↑        & 1.376  & 1.374  & -0.15\% \\
Shannon Entropy ↑    & 10.84  & 10.87  & +0.3\% \\
\bottomrule
\end{tabular}
\end{wraptable}

\subsection{What Makes an Effective Curation Strategy?}

Comparing high-scoring and low-scoring strategies from each category's strategy pool reveals consistent patterns. Effective strategies share four key characteristics: (1) \textit{concrete criteria}—specifying measurable thresholds (e.g., ``over 80\% identical'' for deduplication) rather than vague judgments like ``clearly truncated''; (2) \textit{targeted deletion}—removing specific elements (HTML tags, PII, promotional content) rather than wholesale document filtering; (3) \textit{explicit preservation rules}—protecting domain-critical content alongside deletion instructions, such as ``Preserve theorem statements and proofs'' in mathematics or ``Preserve clinical units and drug names'' in medicine, preventing over-aggressive filtering; and (4) \textit{conservative operations}—preferring deletion of uncertain fragments over risky transformations that may introduce errors.

\subsection{Task-Specific Performance}

To understand where cleaning-focused strategies provide the most benefit, we analyze performance across all 18 benchmarks and identify three distinct patterns, reflecting how different task types respond to data cleaning.

\paragraph{Substantial Improvement (Cleaned $>$ Raw).}

\begin{figure}[htbp]
    \centering
    \begin{subfigure}[b]{0.24\textwidth}
        \includegraphics[width=\textwidth]{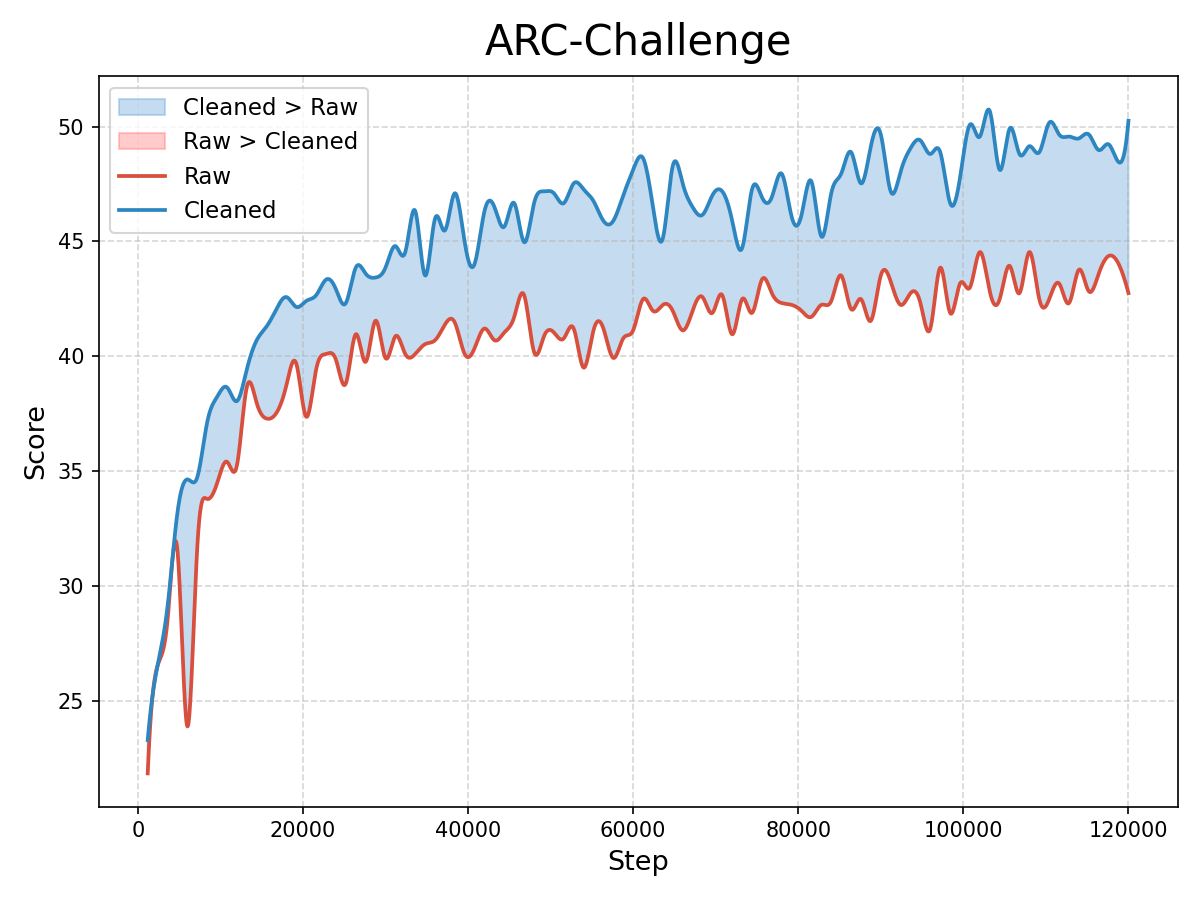}
    \end{subfigure}
    \hfill
    \begin{subfigure}[b]{0.24\textwidth}
        \includegraphics[width=\textwidth]{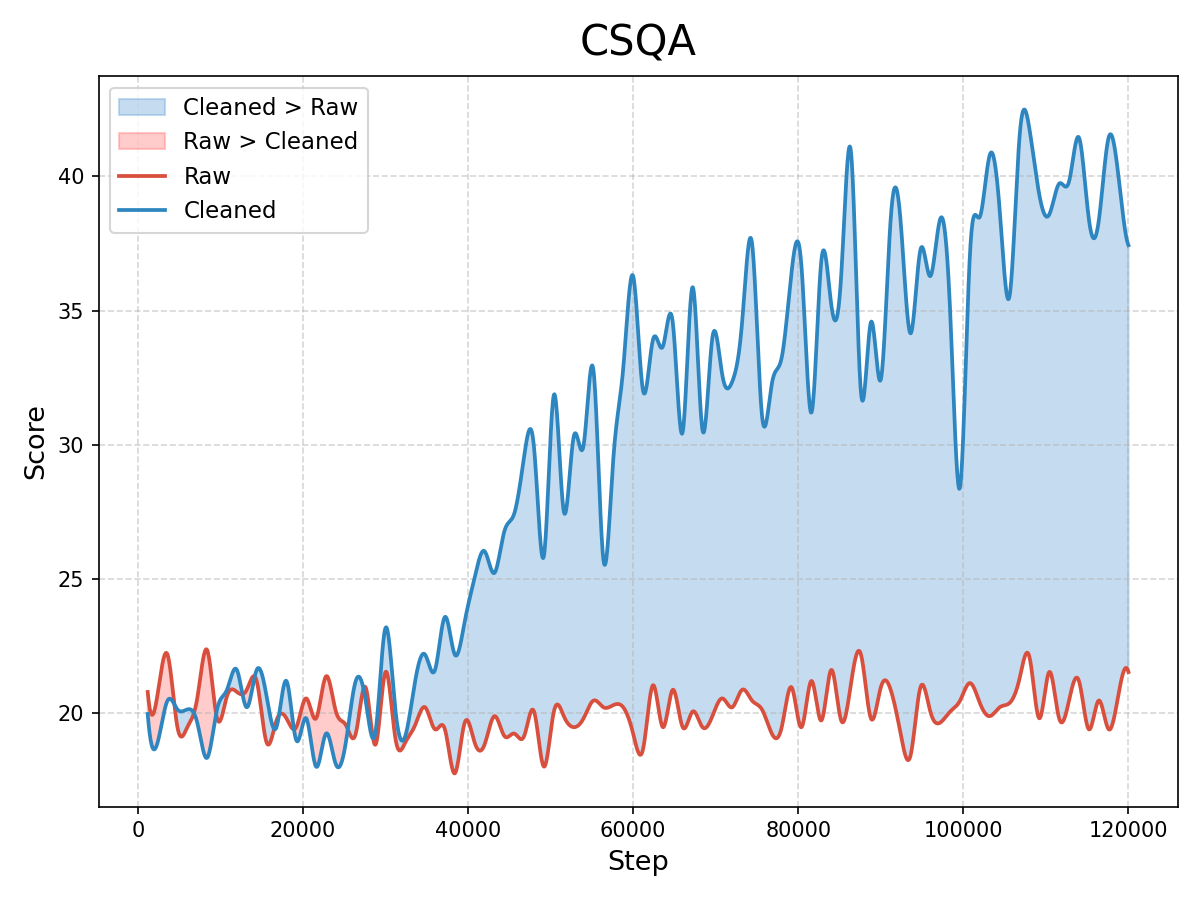}
    \end{subfigure}
    \hfill
    \begin{subfigure}[b]{0.24\textwidth}
        \includegraphics[width=\textwidth]{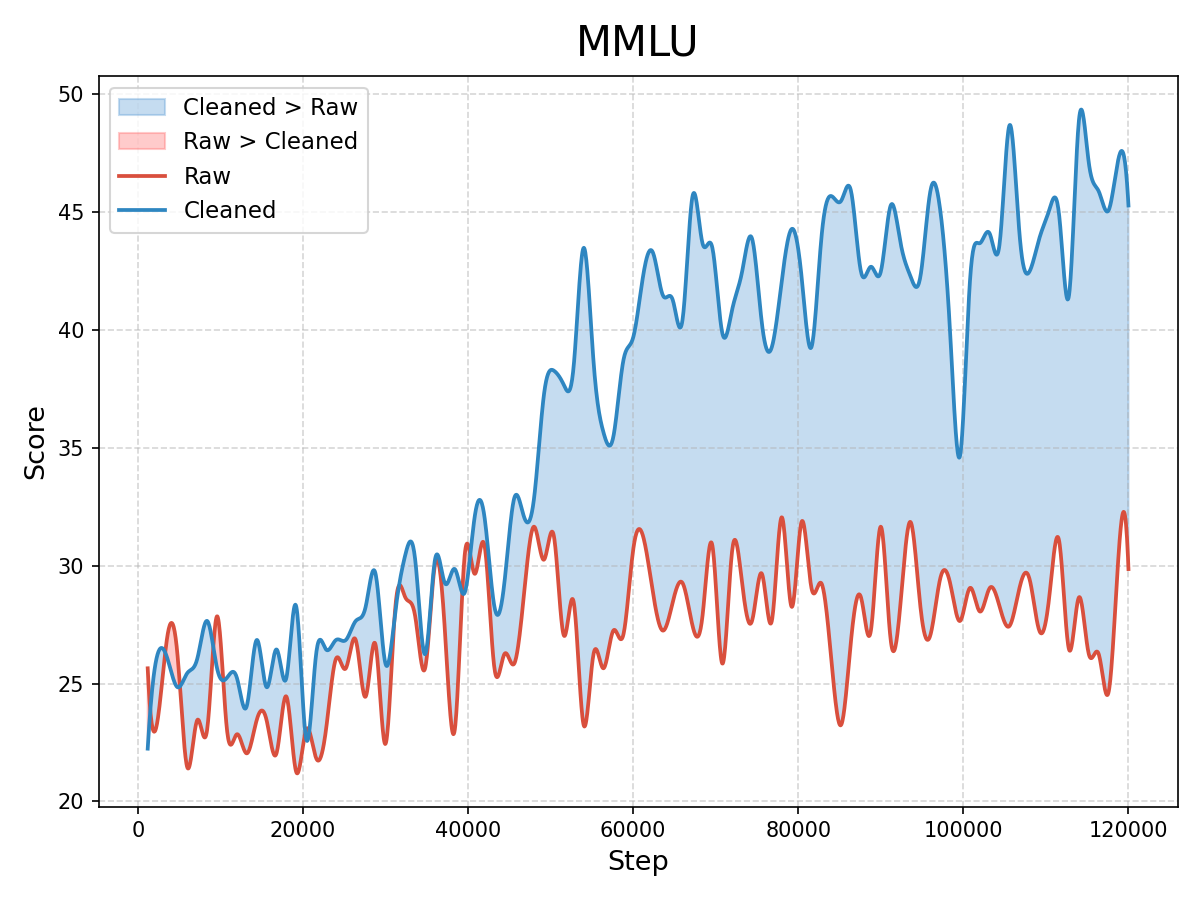}
    \end{subfigure}
    \hfill
    \begin{subfigure}[b]{0.24\textwidth}
        \includegraphics[width=\textwidth]{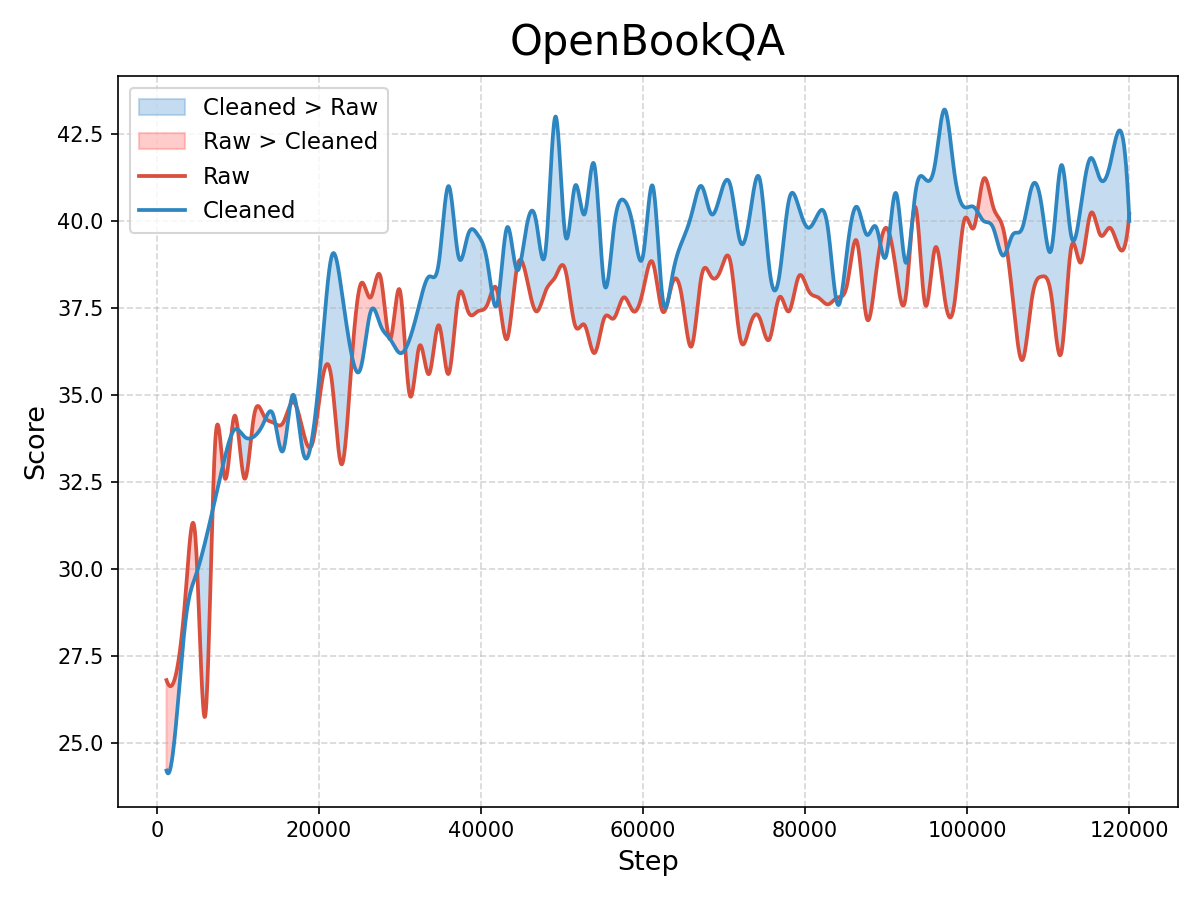}
    \end{subfigure}
    \caption{Representative benchmarks where cleaned data substantially outperforms raw data throughout training. The gap between cleaned (blue) and raw (red) widens consistently as training progresses.}
    \label{fig:clean_higher}
\end{figure}

The most pronounced gains appear on benchmarks centered on knowledge memorization and recall, including MMLU (57 academic subjects across science, law and medicine), CSQA (everyday conceptual knowledge), ARC-Challenge (challenging science questions), OpenBookQA (science facts combined with world knowledge), and medical benchmarks MedQA and MedMCQA. This finding is consistent with concurrent work on guided web rewriting \citep{nguyen2025recycling,karimi2025nemotroncc}, which also observes large MMLU gains when synthetic or cleaned data enriches the factual content available during pretraining. This pattern points to a clear underlying mechanism: raw web documents contain rich factual content obscured by formatting artifacts, HTML noise, and boilerplate text. By removing such surface-level corruption while preserving knowledge-bearing content, \modelname allows models to more effectively memorize and retain factual knowledge during pretraining. As shown in Figure~\ref{fig:clean_higher}, the advantage of cleaned data emerges early and continues to widen throughout training.

\paragraph{Performance Degradation (Cleaned $<$ Raw).}

\begin{figure}[htbp]
    \centering
    \begin{subfigure}[b]{0.24\textwidth}
        \includegraphics[width=\textwidth]{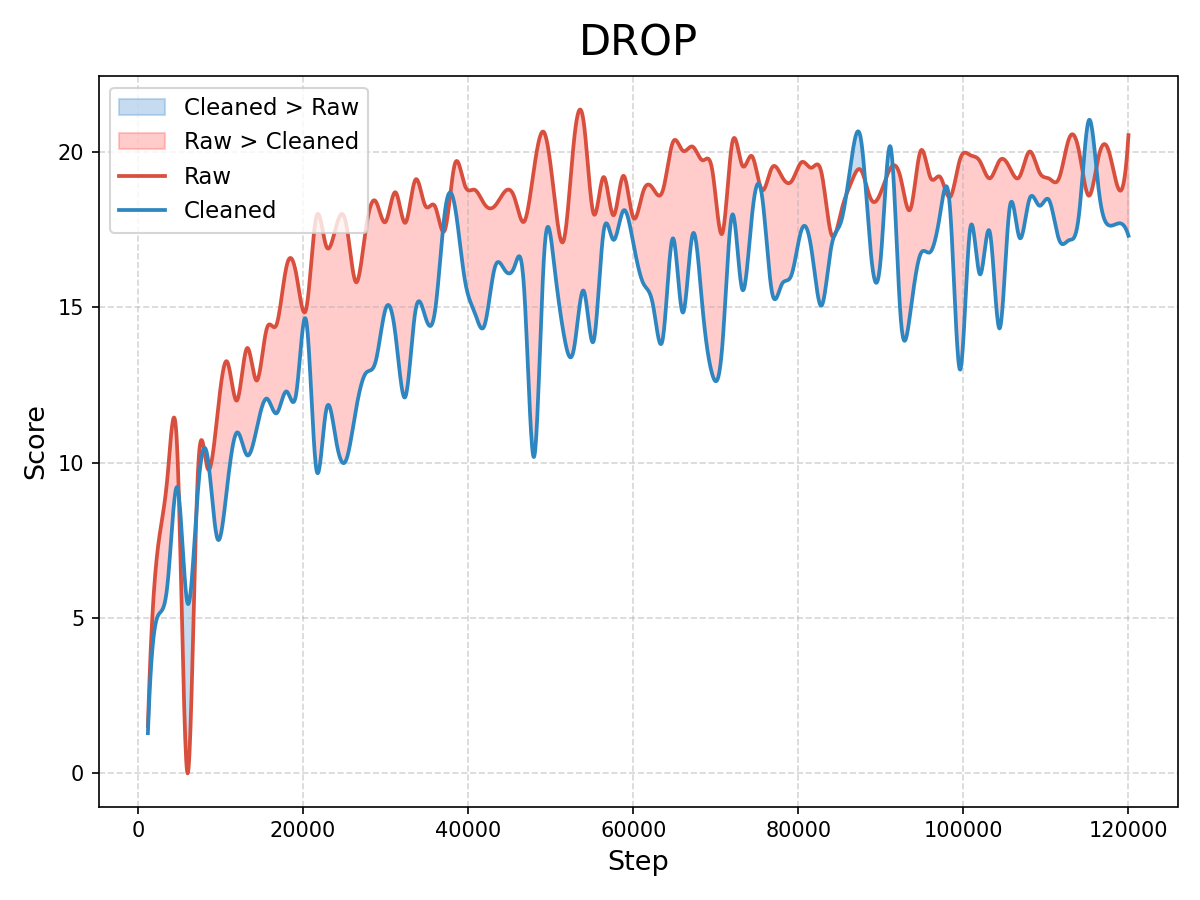}
    \end{subfigure}
    \hfill
    \begin{subfigure}[b]{0.24\textwidth}
        \includegraphics[width=\textwidth]{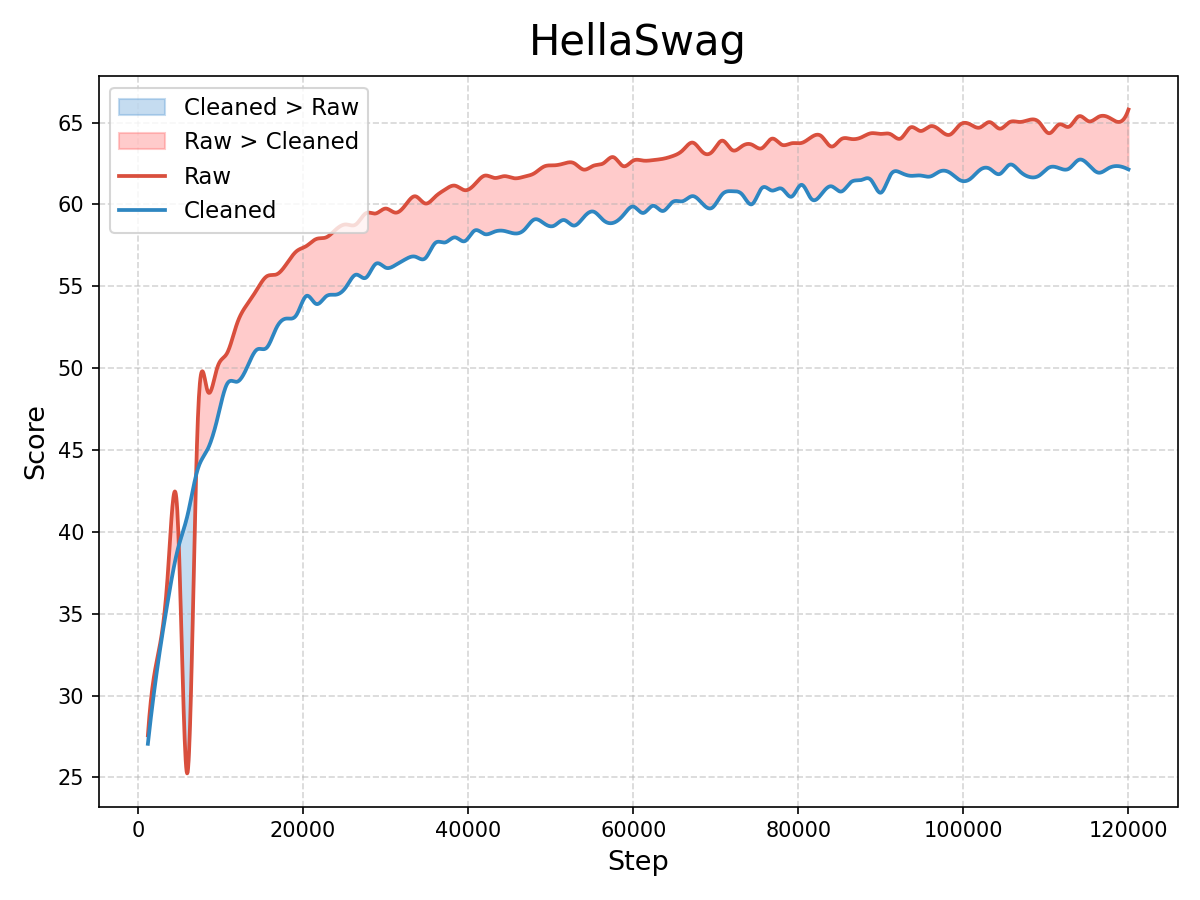}
    \end{subfigure}
    \hfill
    \begin{subfigure}[b]{0.24\textwidth}
        \includegraphics[width=\textwidth]{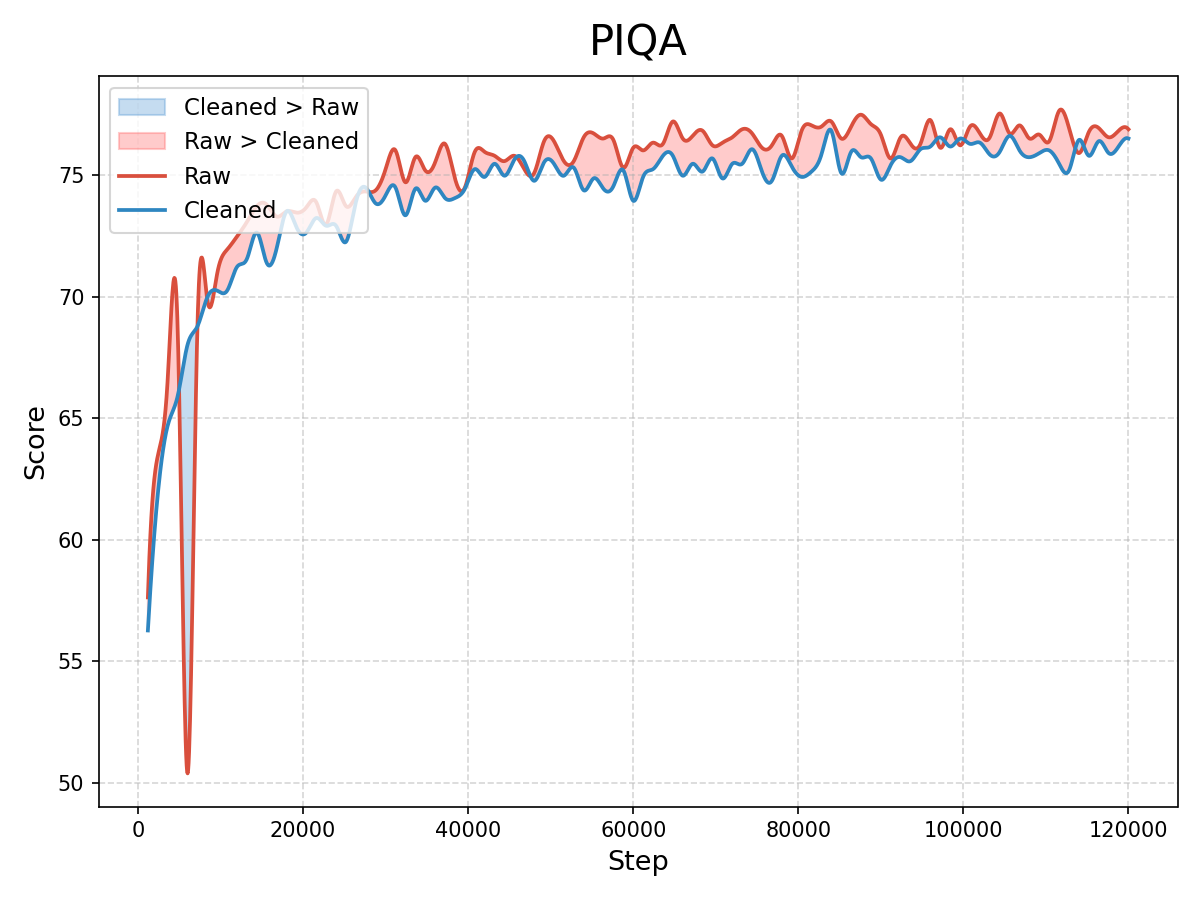}
    \end{subfigure}
    \hfill
    \begin{subfigure}[b]{0.24\textwidth}
        \includegraphics[width=\textwidth]{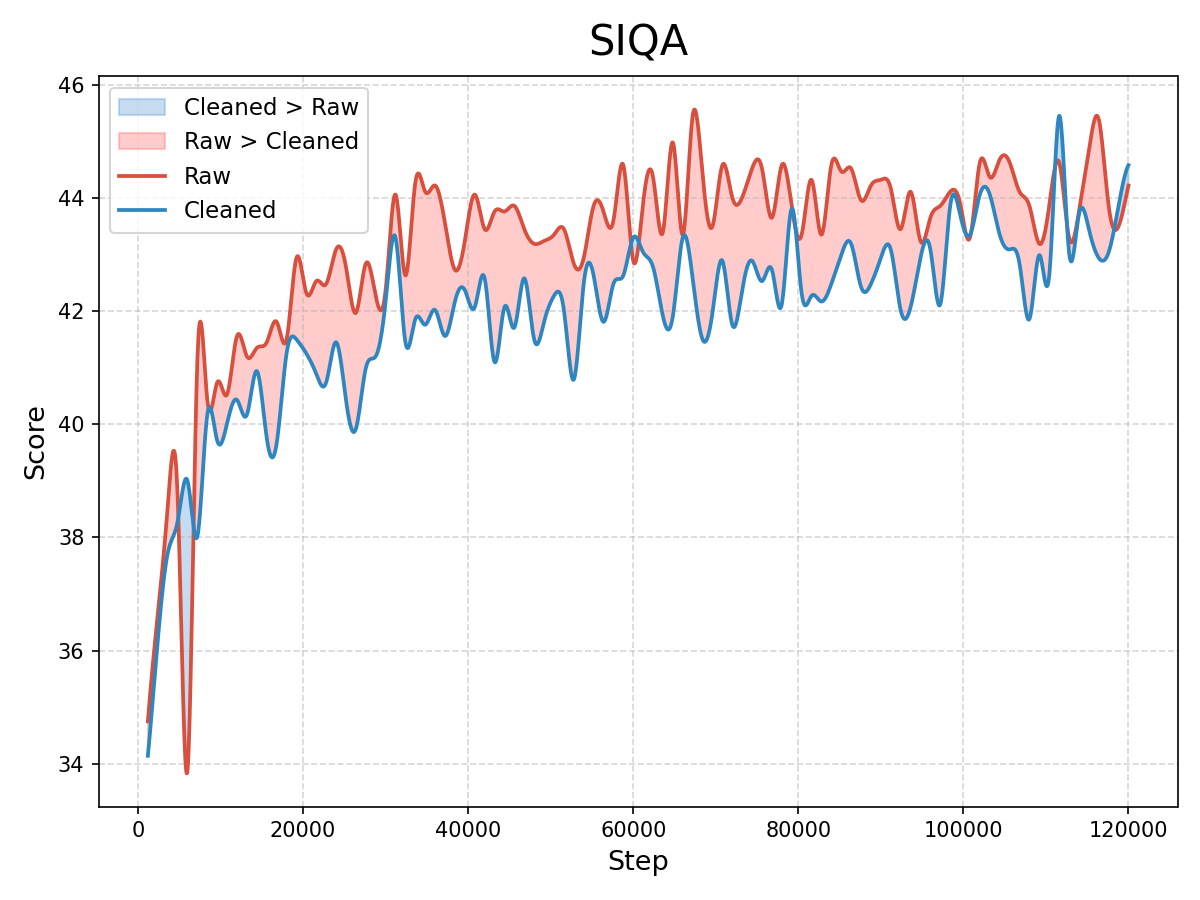}
    \end{subfigure}
    \caption{Representative benchmarks where raw data outperforms cleaned 
    data. The gap between raw (red) and cleaned (blue) 
    is established early and remains stable throughout training.}
    \label{fig:raw_higher}
\end{figure}

A different pattern emerges on benchmarks evaluating informal and situated language understanding, including HellaSwag (plausible continuation of everyday activity descriptions), SIQA (social commonsense reasoning), PIQA (physical commonsense through everyday procedures), and DROP (numerical reasoning over diverse paragraphs). As shown in Figure~\ref{fig:raw_higher}, the performance gap appears from the very beginning of training, suggesting a fundamental difference in what the two data distributions teach the model rather than a matter of convergence speed. We hypothesize that cleaning removes colloquial and conversational expressions from the training corpus, shifting the language distribution toward more formal and structured text. As a result, the model becomes less familiar with the informal registers these benchmarks rely on, producing outputs that are stylistically misaligned with how these tasks naturally present language.

\paragraph{Minimal Change (Cleaned $\approx$ Raw).}

Beyond these two groups, a number of benchmarks show negligible differences between cleaned and raw data, with the two training curves interleaving throughout without a consistent directional gap—representative examples include BBH (multi-step logical and algorithmic reasoning), WinoGrande 
(commonsense reasoning through pronoun resolution), AGIEval (general cognitive abilities from standardized exams), and RACE (reading comprehension grounded in provided passages). Unlike the clear separation seen in the previous two groups, neither data configuration holds a sustained advantage here, suggesting these capabilities are largely robust to the surface-level changes \modelname introduces. One possible explanation is that these tasks depend less on the specific textual properties that cleaning modifies.

\begin{figure}[htbp]
    \centering
    \begin{subfigure}[b]{0.24\textwidth}
        \includegraphics[width=\textwidth]{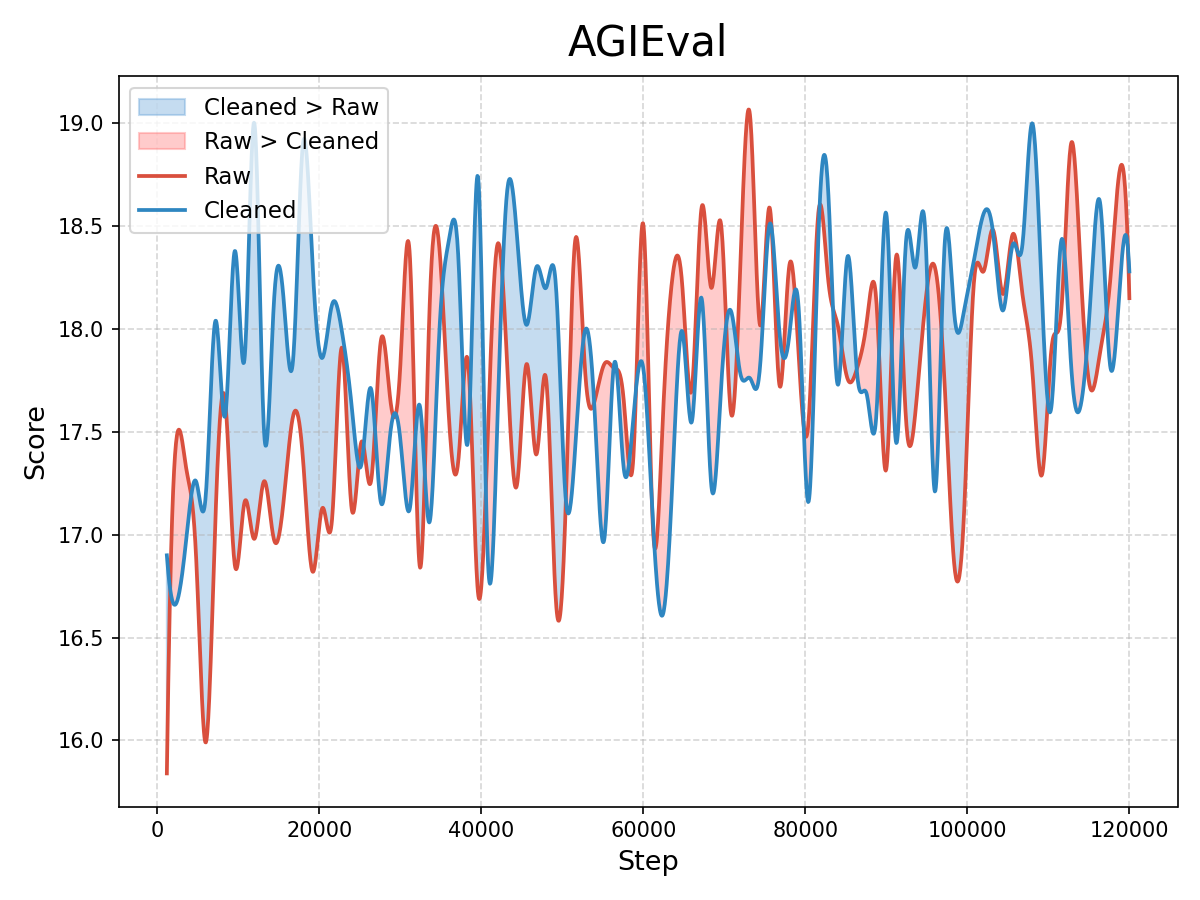}
    \end{subfigure}
    \hfill
    \begin{subfigure}[b]{0.24\textwidth}
        \includegraphics[width=\textwidth]{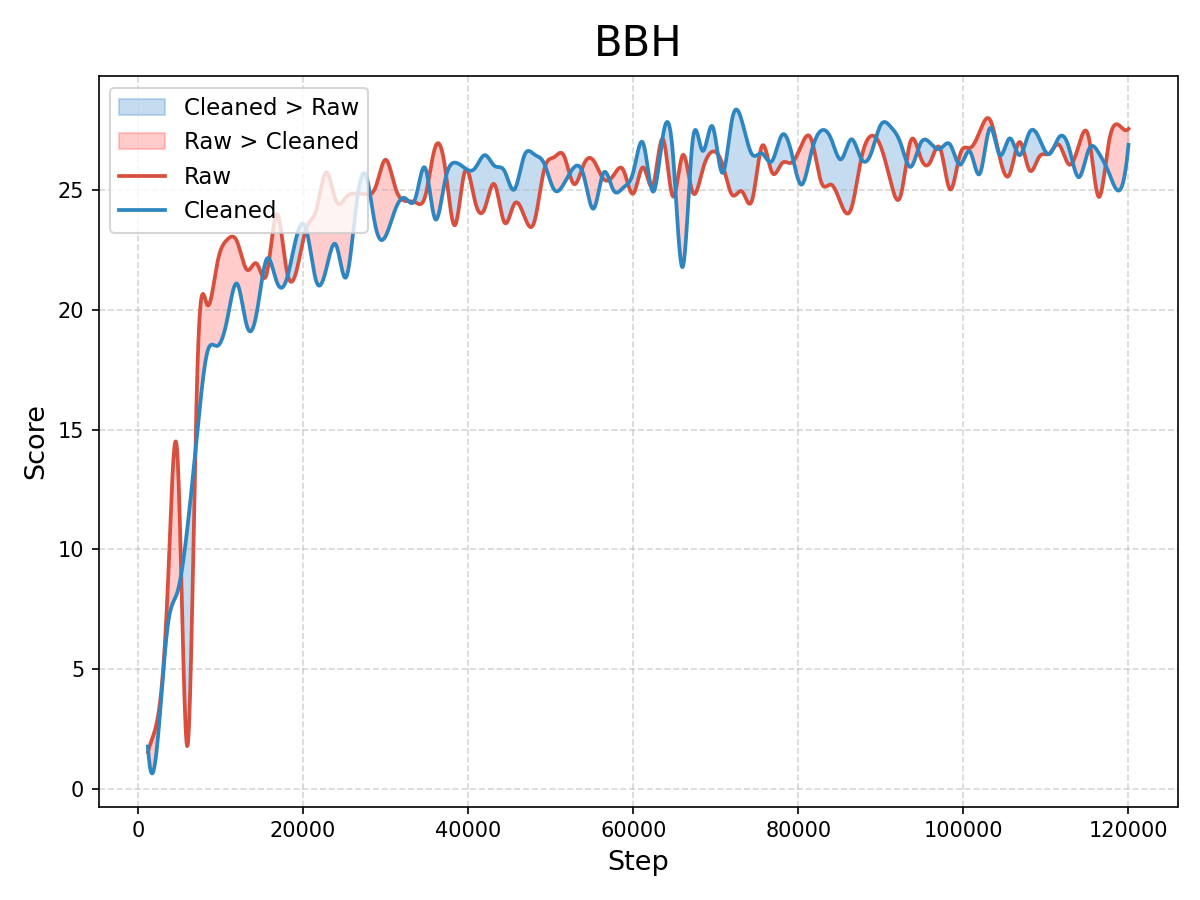}
    \end{subfigure}
    \hfill
    \begin{subfigure}[b]{0.24\textwidth}
        \includegraphics[width=\textwidth]{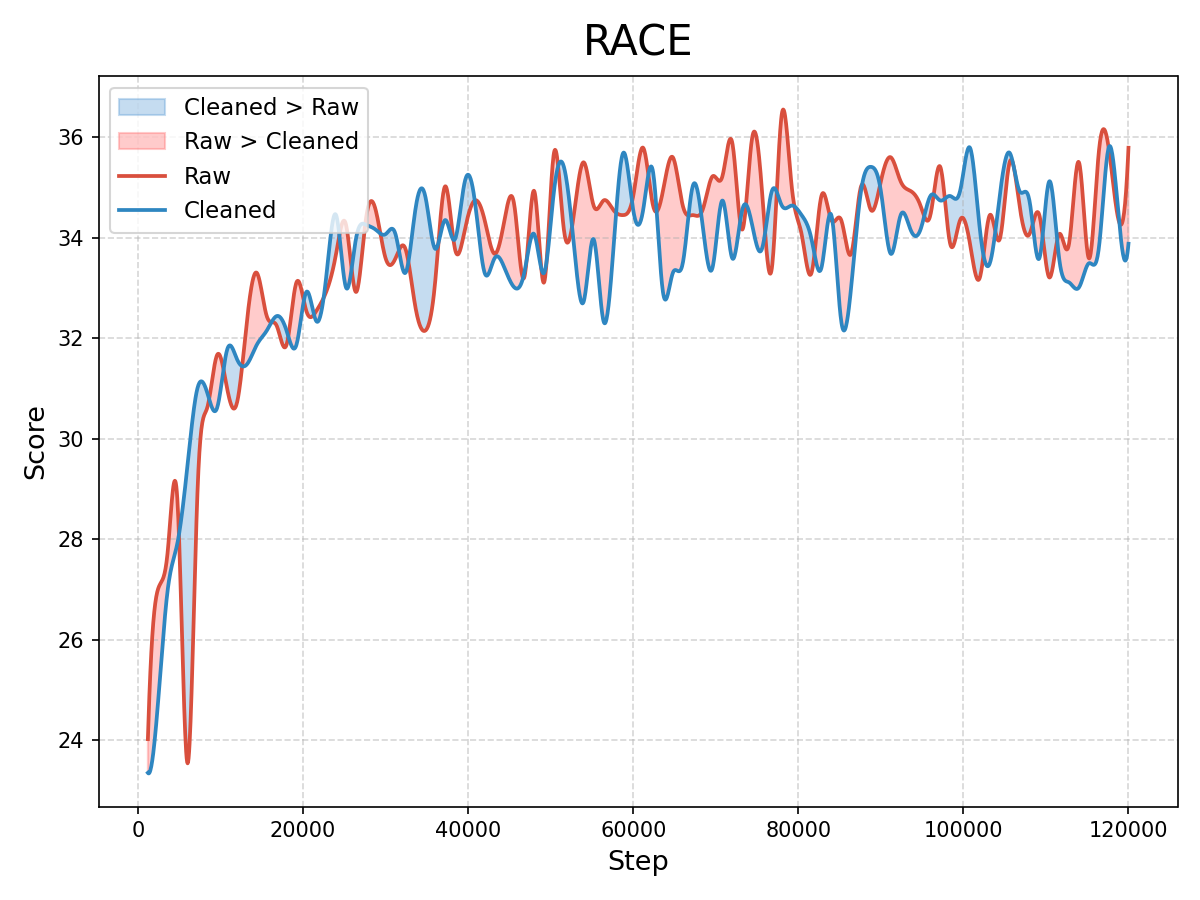}
    \end{subfigure}
    \hfill
    \begin{subfigure}[b]{0.24\textwidth}
        \includegraphics[width=\textwidth]{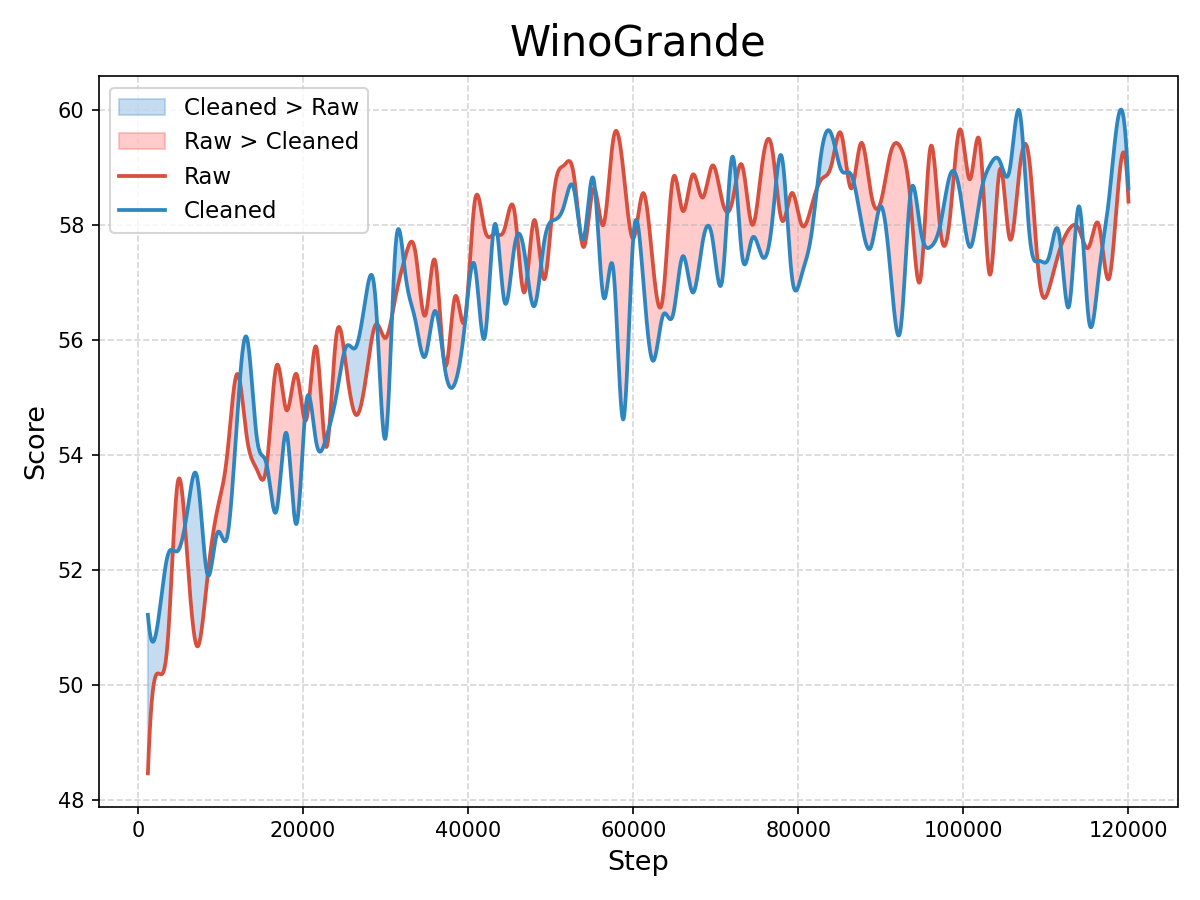}
    \end{subfigure}
    \caption{Representative benchmarks where cleaned and raw data yield comparable performance, with the two curves interleaving throughout training without a consistent directional gap.}
    \label{fig:sim}
\end{figure}
Taken together, these three groups suggest that the impact of data cleaning is not uniform across tasks, and may be mediated by how much a benchmark relies on the specific textual properties that cleaning modifies—such as knowledge density, language register, and surface-level formatting.

\subsection{Semantic Preservation and Data Diversity}
\label{sec:diversity_analysis}

\begin{wrapfigure}{r}{0.45\textwidth}  
\centering
\includegraphics[width=0.48\textwidth]{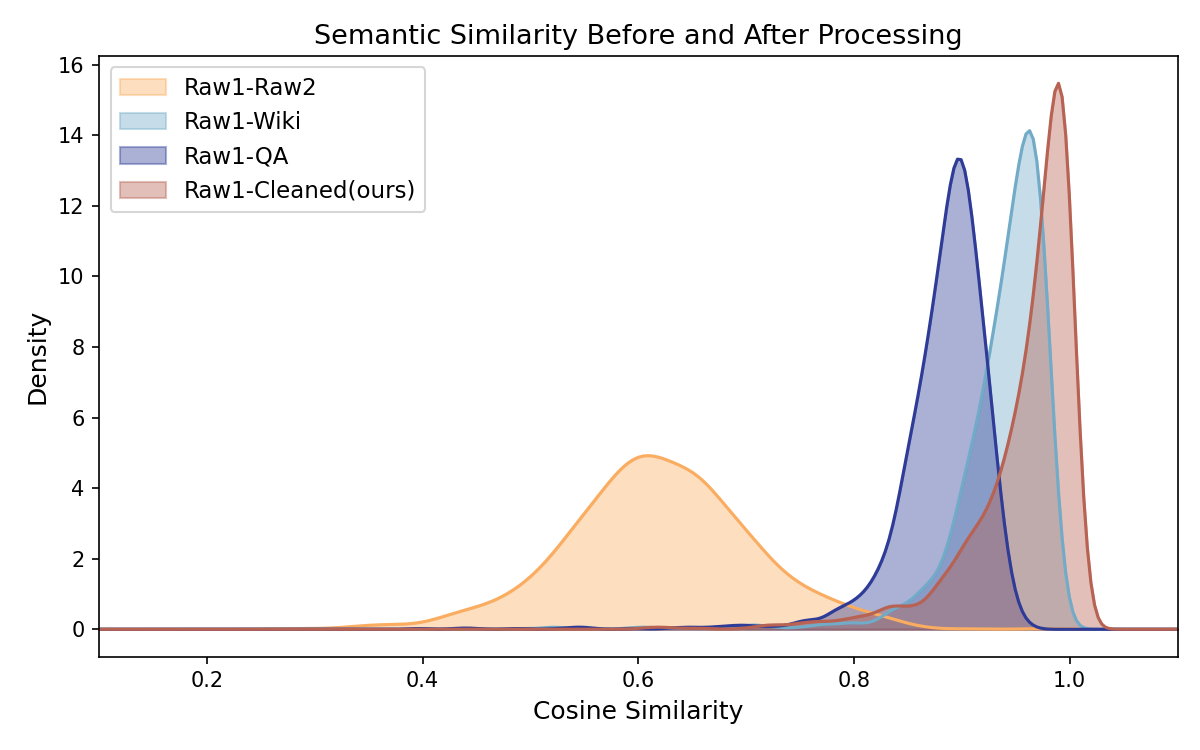}
\caption{Semantic similarity distribution before and after processing. \modelname's cleaning strategies (Raw1-Cleaned) preserve substantially more semantic content than transformation-based methods (Raw1-Wiki, Raw1-QA), with similarity distributions approaching the identity baseline. The tight, right-skewed distribution indicates minimal content modification during cleaning.}
\label{fig:semantic_similarity}
\end{wrapfigure}

\textbf{Semantic preservation.} To validate \modelname's design principle, we analyze how processing affects document semantics. We sample 1,500 documents and compare semantic similarity before and after processing using four methods: \modelname's cleaning strategies, Wikipedia-style rephrasing, question-answer generation, and a random baseline pairing unrelated documents. Figure~\ref{fig:semantic_similarity} reveals that \modelname-cleaned documents maintain substantially higher semantic similarity to their originals (mean: 0.958) compared to transformation-based methods (Wiki: 0.937, QA: 0.882) and far above the random baseline (0.619). The narrow, right-skewed distribution confirms that cleaning operations focus on targeted noise removal rather than content rewriting. This contrasts sharply with methods like question-answer generation, whose broader, left-shifted distribution indicates more aggressive content transformation.

\textbf{Corpus diversity.} While individual documents are minimally modified, does cleaning affect overall corpus diversity? Table~\ref{tab:diversity_metrics} shows that Self-ROUGE-2 (measuring pairwise n-gram overlap) decreases by 21.7\%, indicating documents become more distinct after cleaning—likely due to removal of repeated boilerplate and near-duplicates. Critically, semantic diversity is preserved: mean L2 distance in embedding space remains nearly unchanged (1.376 → 1.374, -0.15\%), and Shannon entropy slightly increases (+0.3\%), suggesting marginally more balanced vocabulary distributions after noise removal.

Together, these results confirm \modelname's dual objective: high per-document semantic similarity (Figure~\ref{fig:semantic_similarity}) demonstrates that cleaning preserves original content, while stable corpus-level diversity metrics (Table~\ref{tab:diversity_metrics}) show that this preservation maintains the broad coverage essential for pretraining.

\section{Related Work}
\label{sec:related_work}

The performance of Large Language Models (LLMs) is fundamentally tied to the quality and scale of their pre-training data. Early efforts focused on curating massive web crawls through heuristic rules and simple classifiers, resulting in datasets like RefinedWeb~\citep{penedo2023refinedweb} and Dolma~\citep{soldaini2024dolma}. Recent advancements have shifted toward more rigorous quality filtering and semantic benchmarking. For instance, DCLM~\citep{li2024datacomplm} formalizes data curation as a standardized competition, while FineWeb-Edu~\citep{penedo2024fineweb} and Ultra-FineWeb~\citep{wang2025ultrafineweb} utilize model-based scoring to extract high-educational-value content. However, these corpora primarily rely on filtering---either retaining or removing documents---which often leads to the loss of valuable domain-specific information that is merely ``noisy'' in its raw form.

Beyond simple filtering, recent studies explore using LLMs as active editors to refine training data. Nemotron-CC ~\citep{karimi2025nemotroncc} and WRAP~\citep{maini2024rephrasing} demonstrate that rephrasing web text into structured formats (e.g., Wikipedia-style or QA) can significantly improve signal density. More fine-grained approaches like ProX~\citep{zhou2024programming} and REFINEX~\citep{bi2025refinex} treat data curation as ``programming every example,'' executing expert-guided edit programs to perform precise corpus surgery. Similarly, Recycling the Web~\citep{nguyen2025recycling} attempts to recover low-quality data through synthetic rephrasing, and GDR \citep{jiang2025generativedatarefinementjust} leverages pretrained models to automatically transform unsafe or private raw content into high-quality training data. While effective, these methods typically require labor-intensive manual design of prompts or programs for each data category, creating a bottleneck when scaling to hundreds of heterogeneous domains.

\section{Conclusion}

This paper presents \modelname, a framework that transforms pretraining data curation from a manual, expertise-intensive process into an automated evolutionary system. By operating a closed loop of data observation, strategy generation, execution, and diagnostic feedback—with knowledge accumulated across generations through an experience pool and a strategy pool—\modelname autonomously discovers high-quality, category-specific cleaning strategies at scale. Applied to 8 categories spanning 672B tokens from Nemotron-CC, the resulting \dataname dataset achieves 44.13 average score across 18 benchmarks, outperforming raw data by 3.96 points and surpassing established corpora including DCLM, Ultra-FineWeb, and FineWeb-Edu, with pronounced gains on knowledge-intensive tasks. Ablation studies further confirm that iterative evolution is essential, demonstrating that only systematically evolved strategies unlock the full latent value of data.

Analysis reveals that strategies evolved independently across diverse categories consistently converge on a cleaning-centric paradigm—targeted noise removal and format normalization with domain-aware content preservation—rather than rewriting text into homogeneous formats. This convergence suggests that systematic cleaning, when guided by properly evolved strategies, suffices for substantial quality improvements without expensive content transformation, offering a simpler and more scalable path for large-scale pretraining data curation. Looking ahead, \modelname establishes evolutionary strategy design as a promising new direction, with future work to explore broader content types, improved stopping criteria, and optimization objectives more directly tied to downstream task performance.

\section{Limitations and Future Directions}
Despite promising results, DataEvolve has several limitations that point to natural directions for future work. First, due to computational constraints, \dataname covers only 8 categories from Nemotron-CC, all of which are academic content in scientific domains. This also means that when comparing \dataname against other general-purpose pretraining corpora, the comparison may not be entirely fair, as those datasets cover a much broader range of topics. We note that our primary contribution is the processed-vs-raw comparison, which directly validates \modelname's effectiveness; the cross-dataset comparison is intended to demonstrate \dataname's value as an open-source resource rather than to claim general superiority. Extending \modelname to broader content types remains an important direction for future work. Second, each category is evolved for a fixed 30 iterations without an adaptive stopping criterion, which may not be sufficient for full convergence; more iterations and principled stopping criteria would likely yield further gains. Finally, a more fundamental challenge lies in our fitness approximation: while sample-based quality scoring enables efficient strategy evolution, it introduces noise relative to true downstream performance. Incorporating lightweight model training as a more direct feedback signal may be a natural next step, though our experiments reveal a subtle difficulty—quality differences between strategies often only emerge after training on hundreds of billions of tokens, and performance rankings can even reverse at earlier stages of training. This suggests that small-scale training signals may themselves be unreliable proxies, and designing an efficient yet accurate evaluation mechanism remains an open challenge for the community.

\bibliographystyle{acl_natbib}
\bibliography{main}
\clearpage
\appendix
\section{Classification}
\label{app:fdc-mapping}
The Dewey Decimal Classification (DDC) is a widely adopted library classification system that systematically organizes knowledge through decimal numerical codes. We merged and remapped the numerical codes from FDC labels to align them with disciplines suitable for current research needs.

\begin{longtable}{lll}
\toprule
\textbf{Higher Level Category } & \textbf{Code Range} & \textbf{Category} \\
\midrule
\endfirsthead
\multicolumn{3}{c}{\tablename\ \thetable{} -- continued} \\
\toprule
\textbf{Higher Level} & \textbf{Code Range} & \textbf{Category} \\
\midrule
\endhead
\bottomrule
\endfoot
\bottomrule
\endlastfoot

\rowcolor{blue!8}
\textbf{computer science} & 000-009 & computer\_science \\
\rowcolor{white}
\textbf{mathematics} & 500-519 & mathematics \\
\rowcolor{blue!8}
\textbf{medicine} & 610-619 & medicine \\
\rowcolor{white}
\textbf{stem-others} & 355-359 & military\_science \\
\rowcolor{white}
 & 520-529 & natural\_sciences\_astronomy \\
\rowcolor{white}
 & 530-539 & physics \\
\rowcolor{white}
 & 540-549 & chemistry \\
\rowcolor{white}
 & 550-559 & natural\_sciences\_earth \\
\rowcolor{white}
 & 560-569 & natural\_sciences\_paleontology \\
\rowcolor{white}
 & 570-579 & biology \\
\rowcolor{white}
 & 580-589 & natural\_sciences\_botany \\
\rowcolor{white}
 & 590-599 & natural\_sciences\_zoology \\
\rowcolor{white}
 & 600-610, 620-621, 626, 629 & engineering \\
\rowcolor{white}
 & 622 & engineering\_mining \\
\rowcolor{white}
 & 623 & engineering\_maritime \\
\rowcolor{white}
 & 624 & engineering\_civil \\
\rowcolor{white}
 & 625 & engineering\_railway \\
\rowcolor{white}
 & 627 & engineering\_water \\
\rowcolor{white}
 & 628 & engineering\_environment \\
\rowcolor{white}
 & 630-631, 632-635, 636-639 & agriculture \\
\rowcolor{white}
 & 660-669 & engineering\_chemical \\
\rowcolor{white}
 & 670-689 & manufacturing \\
\rowcolor{white}
 & 690-699 & construction \\
\rowcolor{white}
 & 910-919 & natural\_sciences\_geography \\
\rowcolor{blue!8}
\textbf{humansocial} & 010-099, 350-354, 640-649, 650-659 & management \\
\rowcolor{blue!8}
 & 100-129, 140-149, 160-199 & philosophy \\
\rowcolor{blue!8}
 & 130-139, 150-159 & psychology \\
\rowcolor{blue!8}
 & 200-299 & religion \\
\rowcolor{blue!8}
 & 300-319, 360-369, 380-399 & sociology \\
\rowcolor{blue!8}
 & 320-329 & political\_science \\
\rowcolor{blue!8}
 & 330-339 & economics \\
\rowcolor{blue!8}
 & 340-349 & law \\
\rowcolor{blue!8}
 & 370-379 & education \\
\rowcolor{blue!8}
 & 400-499 & linguistics \\
\rowcolor{blue!8}
 & 700-709, 750-769 & art\_fine\_arts \\
\rowcolor{blue!8}
 & 710-729 & art\_architecture \\
\rowcolor{blue!8}
 & 730-739 & art\_artifacts \\
\rowcolor{blue!8}
 & 740-749 & art\_design \\
\rowcolor{blue!8}
 & 770-779 & art\_photography \\
\rowcolor{blue!8}
 & 780-789 & art\_music \\
\rowcolor{blue!8}
 & 790-799 & art\_sports \\
\rowcolor{blue!8}
 & 800-899 & literature \\
\rowcolor{blue!8}
 & 900-909, 920-999 & history \\
\end{longtable}

\section{Prompt}


\begin{tcolorbox}[
colback=green!5!white,
colframe=green!35!gray,
  boxrule=2pt,
  arc=8pt,
  title={\textbf{L5 Processing Prompt}},
  fonttitle=\large\bfseries,
  enhanced jigsaw,
  breakable
]
\begin{TextBlock}
You are an expert data quality analyst specializing in identifying issues in text data.

## Your Mission
Observe sample text data and identify quality issues that could negatively impact large language model (LLM) pretraining.

**What is "quality" for pretraining data?**
High-quality pretraining data should:
- **Natural**: Authentic human language, diverse in style and complexity
- **Informative**: Meaningful content, not repetitive filler  
- **Accurate**: Factually sound and logically complete
- **Clean**: Free of noise (HTML, metadata, duplicates, encoding errors)
- **Safe**: Minimal toxicity/PII, preserving legitimate diverse content

## Available Tools
You have access to two tools:

1. **read_experience()**: Read existing quality issues from the experience database
   - Use this to understand what issues have been identified before
   - Avoid duplicating existing issues
   - Build upon previous observations

2. **write_experience(issue: str)**: Write a new quality issue to the experience database
   - Write ONE issue at a time
   - Each issue should be clear, specific, and actionable
   - Call this tool multiple times if you identify multiple issues
   - This is the PRIMARY way to record your findings

## Your Process:
1. First, call `read_experience()` to see what issues have already been identified
2. Analyze the sample data provided in the prompt
3. For each NEW issue you identify, call `write_experience(issue="your specific issue description")`
4. Do NOT duplicate existing issues from the experience database
5. Continue until you've identified all quality issues

## Important:
- Use `write_experience()` tool to actually save issues to the database
- Each call should describe ONE distinct issue
- Be specific and actionable in your descriptions
- Focus on real quality problems that affect model training
- All issues should be saved via tool calls, not just mentioned in text

\end{TextBlock}
\end{tcolorbox}

\begin{tcolorbox}[
colback=blue!5!white,
colframe=blue!30!gray,
boxrule=2pt,
arc=8pt,
title={\textbf{User Prompt for Data Observer}},
fonttitle=\large\bfseries,
enhanced jigsaw,
breakable
]
\begin{TextBlock}
**Raw Data Example {i}:**\n{sample}\n\n"
    
    return f"""# Data Quality Observation Task

## Context
You are analyzing sample data from a dataset that typically contains {config.DOMAIN} content in {config.CONTENT} style. This data will be used for large language model (LLM) pretraining, so identifying quality issues is critical for designing effective cleaning prompts.


## Sample Data to Analyze

Below are {len(samples)} randomly sampled text examples from the dataset:

{sample_text}


## Your Task:

Objectively analyze the samples above and identify ALL quality issues that could negatively impact large language model pre-training, including but not limited to: data noise, incoherence, formatting problems, and any other factors that may degrade training effectiveness.

### Analysis Steps:

1. **First**, call `read_experience()` to review previously identified issues
   - This helps you understand existing patterns and avoid recording duplicates
   - Learn from what has already been discovered

2. **Then**, carefully examine each raw data sample with a critical and independent mindset.
   
   **What to look for**: Anything that would make this data unsuitable for model pretraining.
   
   **Analytical questions to guide your thinking**:
   - If you were training a language model, what problems would this data introduce?
   - What would prevent clear understanding or proper use of this content?
   - What patterns seem inconsistent, incomplete, incorrect, or harmful?
   - What elements don't belong in clean, high-quality training data?
   - Are there domain-specific noise patterns (considering this is {config.DOMAIN} content)?
   
   **Common issue areas** (use as reference only - not exhaustive):
   - Formatting and structure issues
   - Content completeness and coherence
   - Encoding and character problems
   - Noise information
   

3. **For each distinct issue** you identify:
   - Call `write_experience(experience="clear description of the issue based on examples")`
   - Write ONE issue per function call

4. **Finally**, provide a brief summary of your findings:
   - How many distinct issues did you identify?
   - What are the most critical problems?

### Important Guidelines:

- **Quality over quantity**: Focus on real, actionable issues that affect data quality
- **Avoid duplicates**: Check existing experiences first and don't record the same issue twice
- **One issue per call**: Each `write_experience()` should describe ONE distinct problem
- **Think independently**: Don't be biased by the metadata labels - analyze what you actually see
- **Be thorough**: Look for both obvious problems and subtle quality issues

Begin your analysis now.
\end{TextBlock}
\end{tcolorbox}

\begin{tcolorbox}[
colback=green!5!white,
colframe=green!35!gray,
boxrule=2pt,
arc=8pt,
title={\textbf{System Prompt for Stragegy Designer}},
fonttitle=\large\bfseries,
enhanced jigsaw,
breakable
]
\begin{TextBlock}
You are an expert data cleaning prompt designer.

## Available Tools:
- read_experience(): Read all identified data quality issues
- update_cleaning_prompt(prompt_content): Update the cleaning prompt file (must include {text} placeholder)

## Your Role:
Design effective cleaning prompts based on identified quality issues and historical performance. Your prompts guide data cleaning for model training-----never introduce artificial tokens, placeholders, or markers. 

## Core Design Philosophy:
Your primary goal is to preserve the original content's meaning and substance while removing noise and improving quality. Preserve the authenticity of the original document as much as possible. Never introduce placeholders, markers, or artificial tokens (e.g., [REMOVED], [REDACTED], <CLEANED>) into data. If content must be removed, delete it entirely—leave no traces or indicators.

## Design Principles:
1. Clarity: Clear, unambiguous instructions
2. Completeness: Address issues which you think are important from experiences
3. Structure: Logical organization
4. Must include {text} placeholder for the input text

## Your Workflow:
1. **Read Experiences**: Call read_experience() to understand all data quality issues
2. **Review History**: Analyze the historical best prompt and its performance (if provided)
3. **Design New Prompt**: Create a comprehensive cleaning prompt
4. **Save Prompt**: Call update_cleaning_prompt() with your new prompt content
5. **Report Results**: Provide your analysis in the specified format

## Required Output Format:
After completing your work, you MUST output your results in the following JSON format:

```json
{
    "prompt_content": "Your complete cleaning prompt with {text} placeholder",
    "design_rationale": "Explanation of your design decisions and how they address the identified issues",
    "key_improvements": ["List of specific improvements made", "Each improvement should be concise"],
    "expected_benefits": "How this prompt should improve data quality"
}
```

## Important Notes:
- The prompt_content MUST include {text} placeholder exactly as shown
- Call update_cleaning_prompt() tool to actually save your prompt to the file
- Your JSON output should be the last thing in your response
- Ensure the JSON is valid and properly formatted
\end{TextBlock}
\end{tcolorbox}

\begin{tcolorbox}[
colback=blue!5!white,
colframe=blue!30!gray,
boxrule=2pt,
arc=8pt,
title={\textbf{User Prompt for Stragegy Designer}},
fonttitle=\large\bfseries,
enhanced jigsaw,
breakable
]
\begin{TextBlock}
    if best_prompt:
        historical_section = f"""
## Historical Best Prompt

You have access to the best-performing prompt from previous iterations:

### Previous Prompt Content
```
{best_prompt.get('prompt_content', 'N/A')}
```

### Previous Prompt Cleaning Quality Analysis
{best_prompt.get('clean_analysis', 'No analysis available')}


### Your Task with Historical Context
1. **Analyze what worked**: Identify strengths in the previous prompt
2. **Identify gaps**: Find issues not addressed or poorly handled
3. **Improve iteratively**: Build upon successes, fix weaknesses based on the analysis of previous prompt
4. **Maintain continuity**: Don't discard effective strategies

**Important**: This is an evolution, not a revolution. Preserve what works while addressing identified problems.
"""
    else:
        historical_section = """
## First Prompt Design

This is the first prompt for this dataset. No historical performance data is available.

### Your Task
1. **Understand the issues**: Read the experience database thoroughly
2. **Design comprehensively**: Address all major quality issues
3. **Be specific**: Provide clear, actionable instructions
4. **Set foundation**: Create a solid baseline for future iterations



**Important**: This prompt will serve as the foundation for future improvements, so make it comprehensive and well-structured.
"""
    
    return f"""# Data Cleaning Prompt Design Task

**What you'll likely see in these samples**:
- Most content is related to: {config.DOMAIN}
- Common content style: {config.CONTENT}

{historical_section}

## Your Workflow:

1. **Read Experiences**: Call read_experience() to see all data quality issues
2. **Review History**: Analyze the historical prompt and its performance (if provided)
3. **Design Prompt**: 
   - Address specific issues documented in the experience database
   - **Identify domain-specific quality patterns**:
     * Think deeply about the content characteristics of {config.DOMAIN}
     * What quality issues are common in this content? (formatting, terminology, structure, etc.)
     * What valuable content should be preserved in this domain?
     * What noise patterns are particularly prevalent in this domain?
   - **Important**: While the dataset is predominantly {config.DOMAIN} content, do not simply remove all non-domain text. The focus is on identifying and handling quality issue patterns **specific to this domain**, not enforcing domain filtering
   - Include {{text}} placeholder
   - Use clear, specific, concise instructions
   - Balance specificity with generalization
4. **Update**: Call update_cleaning_prompt() with your new prompt
5. **Explain**: Provide rationale, improvements, and expected benefits

**Never introduce placeholders or markers that don't exist in natural human text:**
- Forbidden: [REMOVED], [REDACTED], [EMAIL], <CLEANED>, etc.
- Correct: Delete completely, no traces

Artificial tokens pollute training data distribution and confuse the model.

Begin now.


\end{TextBlock}
\end{tcolorbox}

\begin{tcolorbox}[
colback=green!5!white,
colframe=green!35!gray,
boxrule=2pt,
arc=8pt,
title={\textbf{System Prompt for Quality Judge}},
fonttitle=\large\bfseries,
enhanced jigsaw,
breakable
]
\begin{TextBlock}
You are an expert data quality evaluator specializing in assessing data cleaning prompt effectiveness.

## Your Mission
Evaluate the quality of cleaned data and analyze whether the cleaning prompt has good coverage and executability.

## Available Tools

**read_experience()**
- Loads known quality issues from the experience database
- Must be called at the beginning to understand what issues the prompt should address
- Provides context for evaluating coverage

**write_experience(issue: str)**
- Documents NEW issues discovered during evaluation
- Call once per new issue found
- Use when you identify problems not yet in the database

## Evaluation Standards

### Data Pair Scoring (1-10 scale)

Evaluate each data pair on three dimensions:

**1. Cleaning Completeness**
- Are the targeted issues fully resolved?
- Do any original problems remain in the cleaned version?
- Were all necessary cleaning operations performed?

**2. Quality Improvement**
- Is the cleaned text noticeably better than the original?
- Is it more readable?
- Does cleaning add value to the data?

**3. Output Quality**
- Is the semantic meaning preserved?
- Are there unnecessary large deletions of substantive content compared to the original?
- Does the cleaned text introduce confusing or misleading elements?


**Score Guidelines:**
- **9-10 (Excellent)**: All issues cleaned, significant quality improvement, high output quality
- **7-8 (Good)**: Major issues resolved, clear improvement, minor acceptable flaws
- **5-6 (Acceptable)**: Partial cleaning, limited improvement, noticeable problems remain
- **3-4 (Poor)**: Many unresolved issues, minimal or no improvement, quality concerns
- **1-2 (Very Poor)**: Cleaning failure, quality degraded, serious problems introduced

### Prompt Analysis Framework

Analyze the cleaning prompt on two critical dimensions:

**Coverage Analysis**
- What types of issues does the prompt explicitly address?
- What issues are missing or not covered by the prompt?
- How well does the prompt align with known issues in the experience database?
- Are there gaps between what should be cleaned and what the prompt instructs?

**Executability Analysis**
- Which instructions are clear, specific, and unambiguous?
- Which instructions are vague, unclear, or open to interpretation?
- Where does execution fail or vary unexpectedly?
- Are instruction boundaries well-defined (e.g., when to apply vs. not apply)?

### Improvement Suggestions

Provide specific, actionable recommendations on how to modify or improve the cleaning prompt to address the identified issues.(such as:adding instructions for uncovered issues,clarifying ambiguous instructions and so on)

## Output Format

Return a valid JSON object with this structure:
```json
{
  "pair_evaluations": [
    {
      "pair_id": 1,
      "score": 8.5,
      "comment": "Brief explanation of score covering what was done well and what issues remain"
    }
  ],
  "prompt_analysis": "Comprehensive and concise text analysis including:
    - Coverage: issues addressed and missing
    - Executability: clear vs ambiguous instructions, consistency
    - Suggestions: prioritized, specific improvements"
}
```

## General Guidelines

- **Be Objective**: Base scores on clear evidence from the data pairs
- **Be Actionable**: Provide clear, implementable suggestions
- **Be Balanced**: Acknowledge both strengths and weaknesses

\end{TextBlock}
\end{tcolorbox}

\begin{tcolorbox}[
colback=blue!5!white,
colframe=blue!30!gray,
boxrule=2pt,
arc=8pt,
title={\textbf{User Prompt for Quality Judge}},
fonttitle=\large\bfseries,
enhanced jigsaw,
breakable
]
\begin{TextBlock}
for i, pair in enumerate(data_pairs, 1):
        pairs_text += f"""
### Data Pair {i}

**Original Text:**
```
{pair['text']}
```

**Cleaned Text:**
```
{pair['refined_text']}
```

---
"""
    
    return f"""# Data Cleaning Quality Evaluation - Iteration {iteration}


## Cleaning Prompt Under Evaluation

The following prompt was used to clean the data:

---
{prompt_content}
---

## Your Task

### Phase 1: Context Loading
**Call read_experience()** to load the known quality issues from the experience database.

This provides essential context for:
- Understanding what issues this prompt was designed to address
- Evaluating whether the prompt has adequate coverage
- Identifying gaps between known issues and prompt instructions

### Phase 2: Data Pair Evaluation

Evaluate each of the {len(data_pairs)} data pairs below.

For each pair, compare the original and cleaned versions:
- Assess cleaning completeness, quality improvement, and output quality
- Assign a score from 1-10 based on the evaluation standards
- Write a concise comment explaining your score

Focus on:
- Were the issues that should be cleaned actually cleaned?
- Is the cleaned version better than the original?
- Is the output coherent and free of artifacts?

### Phase 3: Prompt Analysis

Analyze the cleaning prompt itself:

**Coverage Perspective:**
- Cross-reference the prompt instructions with issues from the experience database
- Identify which issue types are addressed and which are missing
- Note any new issue types discovered in the data pairs

**Executability Perspective:**
- Examine each instruction in the prompt for clarity and specificity
- Check if the cleaning results show consistent execution across samples
- Identify instructions that are too vague or ambiguous
- Note where the cleaning model failed to follow instructions properly

### Phase 4: Documentation and Recommendations

**If you discover new issues** not present in the experience database:
- Call write_experience(issue) to document each new issue
- Provide a clear description of the issue

**Provide improvement suggestions:**
- Prioritize by impact: HIGH, MEDIUM, or LOW
- Classify by type: Coverage or Executability
- Give specific, actionable recommendations (e.g., exact instruction to add)

## Data Pairs

{pairs_text}

## Expected Output Format

Return a JSON object following this structure:

{{
  "pair_evaluations": [
    {{
      "pair_id": 1,
      "score": 7,
      "comment": "Brief explanation of score covering what was done well and what issues remain"
    }},
    {{
      "pair_id": 2,
      "score": 8.5,
      "comment": "Brief explanation of score covering what was done well and what issues remain"
    }}
  ],
  "prompt_analysis": "Comprehensive but concise text analysis including: Coverage (issues addressed and missing), Executability (clear vs ambiguous instructions, consistency), and prioritized specific improvement suggestions."
}}  

**Important Notes:**
- Output MUST be valid JSON only
- Provide scores for all {len(data_pairs)} pairs in order
- Start by calling read_experience()
- Be specific and actionable in your analysis

Begin your evaluation now.
\end{TextBlock}
\end{tcolorbox}

\section{Case}
\begin{CompareBox}[Example 1 ( High-Medicine-Academic Final Prompt)]{cyan}
\small
Clean the following medical or clinical academic text while preserving only the content explicitly present. Output must be plain UTF-8 text only; do not insert any markup, rich-text tokens, placeholders, or formatting markers. Delete all unwanted material completely, without leaving traces.

Input:
{text}

Instructions:
1. Do not introduce any new clinical facts, procedures, device names, drug names (brand or generic), dosages, methods, or qualifiers not present in the source. If the source text is ambiguous or incomplete, delete the ambiguous clause rather than inferring or adding details.

2. Remove all promotional or marketing language, calls to action, product endorsements, vendor names, affiliate links, and platform branding. If a brand name is not essential to clinical meaning, replace it with a neutral descriptor (e.g., "commercial assay").

3. Strip all user-interface and site metadata, including headers, footers, slide or page numbers, navigation labels, view counts, upload dates, bylines, course or module tags, and other extraneous text not part of the core clinical narrative.

4. Delete all personal identifiers: names, titles, emails, phone numbers, addresses, usernames, institutional bylines, and sign-offs. If an institutional affiliation is essential for context, replace it with a neutral phrase (e.g., "multiple academic centers").

5. Remove HTML tags, escaped entities, inline code, style attributes, emojis, emoticons, and any non-linguistic symbols or stray control characters.

6. Delete references to missing assets (figures, tables, images) and placeholder markers. If tabular data appears intact, serialize it as plain-text label–value pairs (one per line) without reconstructing missing cells.

7. Remove orphaned citation markers (e.g., "[1]", "(Smith et al.)") if no matching reference entry exists. If a reference list is included, preserve full entries and retain only corresponding in-text citations.

8. For unsupported or unreferenced medical claims or numeric values, either apply precise hedging (e.g., "has been reported", "may") if the source text contains the complete claim, or delete the clause entirely. Do not retain large quantitative statements without explicit attribution.

9. Collapse exact or near-verbatim duplicates by keeping the most complete version. Do not merge content across versions if it risks introducing new information.

10. Correct obvious typos, split merged tokens, normalize spacing, and restore sentence boundaries without altering domain-specific terminology.

11. Preserve clinical units and chemical or ionic notation exactly as presented; do not infer or append missing units or convert values.

Output:

Return only the cleaned, high-quality academic medical text.
\end{CompareBox}

\begin{CompareBox}[Example 1( Text Before Clean)]{red}
\small
Three main sources of the disease have been identified in Africa: Lake Tanganyika in the west, Lake Kivu also in the west and Lake Katebe in the more southerly region of the Democratic Republic of Congo. To avoid any risk of infection the rule is simple – don't bathe or walk in bare feet in fresh water. Particularly in Africa and Asia! Online press service recognized by the CPPAP (Joint Committee for Publications and Press Agencies) under no. 0318 W 90885 (Article 1 of Act 86-187 of 1 August 1986 reforming the legal system governing the press).
\end{CompareBox}

\begin{CompareBox}[Example 1( Text After Clean)]{green}
\small
Three main sources of the disease have been identified in Africa: Lake Tanganyika in the west, Lake Kivu also in the west and Lake Katebe in the more southerly region of the Democratic Republic of Congo. Avoid bathing or walking barefoot in fresh water, particularly in Africa and Asia.
\end{CompareBox}

\begin{CompareBox}[Example 2( NotHigh-Mathematics-Academic Final Prompt)]{cyan}
\small
Clean the following text:

{text}

Perform the following steps in order:
1. Remove all UI or web boilerplate, navigation labels (e.g., "Related Posts", "Subscribe"), cookie/privacy banners, raw URLs, site branding, affiliate links, viewcounts, marketing calls-to-action, pagination markers, feed metadata, HTML/XML tags, and scraping artifacts.

2. Delete promotional or marketing content (ads, subscription prompts, service pitches, product descriptions) and remove event schedules, museum or seminar announcements, community notices, and local contact information.

3. Remove all personally identifiable contact information: postal addresses, phone numbers, email addresses, and social media handles. Delete completely; do not insert placeholders. Retain non-contact bibliographic author names and publication citations when present.

4. Identify and preserve all mathematical content: definitions, theorems, proofs, problem statements, solution steps, conceptual explanations, worked examples, and applied-math descriptions. Remove any material whose primary intent is not mathematical reasoning or pedagogy.

5. Collapse exact duplicates and trivially near-duplicate sentences (differing only by punctuation or common stopwords), while retaining distinct methods or solutions with substantive differences.

6. Remove conversational or informal register: filler words (e.g., "um", "like"), greetings, slang, emojis, chat transcripts, and author-aside commentary. Maintain an academic, formal tone.

7. Drop incomplete or unrecoverable fragments (orphaned headings, lists, captions, or tables without context). If a fragment can be fixed through minimal unambiguous syntax repair, do so; otherwise delete it.

8. Normalize punctuation, whitespace, and capitalization: correct spacing around punctuation, remove duplicate spaces, ensure clear sentence boundaries, and apply consistent capitalization.

9. Normalize all mathematical expressions to LaTeX: use \texttt{\$...\$} for inline math and \texttt{\$\$...\$\$} for display math, converting Unicode fractions, symbols, and plain ASCII math accordingly. Do not use \texttt{\textbackslash[\textbackslash]} or other delimiters. Preserve existing well-formed LaTeX, repairing only stray artifacts (unmatched \texttt{\textbackslash left/\textbackslash right} or delimiters). When converting units, maintain a single-unit representation; do not produce ambiguous mixed-unit expressions.

10. Do not alter substantive mathematical logic or correctness. Correct only unambiguous OCR or transcription errors (merged/split digits or clear glyph swaps). If uncertain, leave the original token unchanged.

11. Present essential code in a plain fenced code block (```language```). Remove inline compiler prompts and execution output. If code is not essential to the mathematics, delete it.

12. Convert well-formed tables to plain ASCII tables with consistent columns. Remove malformed tables or tables whose conversion would require guessing content.

Return only the cleaned content. Do not include the cleaning instructions, audit logs, editorial notes, or any internal markers.
\end{CompareBox}

\begin{CompareBox}[Example 2( Text Before Clean)]{red}
\small
2 Answers\\2\\\\We have $f_n(x)=n^2x(1-x^2)^n$ and $f(x)=0$. We need to show that\\$$\lim_{n\to \infty}\sup_{x\in [0,1]}\left|f_n(x)\right|\neq 0$$\\Because $f$ is continuous $$M_n=\sup_{x\in [0,1]}\left|f_n(x)\right|=\max_{x\in [0,1]}\left|f_n(x)\right|=\max_{x\in [0,1]}f_n(x)$$\\To calculate this observe that for $0<x<1$,\\$$f_n^{\prime}(x)=0=\Leftrightarrow 1-x^2-2nx^2=0\Leftrightarrow x=\frac{1}{\sqrt{2n+1}}$$\\Therefore,\\$$M_n=\max\left\{f_n(0),f_n(1),f_n(\frac{1}{\sqrt{2n+1}})\right\}=\frac{n^2}{\sqrt{2n+1}}(1+\frac{1}{2n})^{-n}$$\\Obviously, $M_n\to +\infty$\\\\HINT: derivative of $f_n(x)$, and you will get $$ n^2(1-x^2)^n - n^3x^2(1-x^2)^{n-1}$$ Now notice last expresion is $0$ iff $x = +- \frac{1}{\sqrt{n+1}} $\\Now, plug this into $f_n$ to see where the sup of this functions is: This would be your $M_n$.
\end{CompareBox}

\begin{CompareBox}[Example 2( Text After Clean)]{green}
\small
We have $f_n(x)=n^2x(1-x^2)^n$ and $f(x)=0$. We need to show that\\$$\lim_{n\to\infty}\sup_{x\in[0,1]}\left|f_n(x)\right|\neq0.$$\\Because $f$ is continuous,\\$$M_n=\sup_{x\in[0,1]}\left|f_n(x)\right|=\max_{x\in[0,1]}\left|f_n(x)\right|=\max_{x\in[0,1]}f_n(x).$$\\To calculate this observe that for $0<x<1$,\\$$f_n'(x)=0\;\Longleftrightarrow\;1-x^2-2nx^2=0\;\Longleftrightarrow\;x=\frac{1}{\sqrt{2n+1}}.$$\\Therefore,\\$$M_n=\max\{f_n(0),\,f_n(1),\,f_n\!\left(\frac{1}{\sqrt{2n+1}}\right)\}\\=\frac{n^2}{\sqrt{2n+1}}\left(1+\frac{1}{2n}\right)^{-n},$$\\and clearly $M_n\to+\infty$.\\\\HINT: Differentiate $f_n(x)$; you obtain\\$$n^2(1-x^2)^n - n^3x^2(1-x^2)^{\,n-1}.$$\\This expression is zero iff $x=\pm\frac{1}{\sqrt{n+1}}$.\\Plug this into $f_n$ to see where the supremum of the functions occurs; this gives $M_n\,. $
\end{CompareBox}
\end{document}